\documentclass[10pt,twocolumn,letterpaper]{article}

\usepackage{cvpr}

\definecolor{cvprblue}{rgb}{0.21,0.49,0.74}
\usepackage[pagebackref,breaklinks,colorlinks,allcolors=cvprblue]{hyperref}
\usepackage{amsmath}
\usepackage{algorithm}
\usepackage{algorithmicx}
\usepackage{algpseudocode}
\usepackage{array}
\usepackage{graphicx}

\def\paperID{8} 
\def\confName{CVPR}
\def\confYear{2026}

\title{DocRevive: A Unified Pipeline for Document Text Restoration}

\author{Kunal Purkayastha$^1$, Ayan Banerjee$^1$, Josep Lladós$^1$, Umapada Pal$^2$ \\
$^1$\textit{Computer Vision Center (\{kunal,abanerjee,josep\}@cvc.uab.es)}, \\
$^2$\textit{Indian Statistical Institute, Kolkata (umapada@isical.ac.in)} \\}
\newcommand{\myparagraph}[1]{\vspace{0pt}\noindent{\bf #1}}

\begin{document}

\maketitle              
\begin{abstract}
    In Document Understanding, the challenge of reconstructing damaged, occluded, or incomplete text remains a critical yet unexplored problem. Subsequent document understanding tasks can benefit from a document reconstruction process. In response, this paper presents a novel unified pipeline 
    combining state-of-the-art Optical Character Recognition (OCR), advanced image analysis, masked language modeling, and diffusion-based models to restore and reconstruct text while preserving visual integrity. 
    We create a synthetic dataset of 30{,}078 degraded document images that simulates diverse document degradation scenarios, setting a benchmark for restoration tasks. 
    Our pipeline 
    detects and recognizes text, identifies degradation with an occlusion detector, and uses an inpainting model for semantically coherent reconstruction. A diffusion-based module seamlessly reintegrates text, matching font, size, and alignment. To evaluate restoration quality, we propose a Unified Context Similarity Metric (UCSM), incorporating edit, semantic, and length similarities with a contextual predictability measure that penalizes deviations when the correct text is contextually obvious. Our work advances document restoration, benefiting archival research and digital preservation while setting a new standard for text reconstruction. The OPRB dataset and code are available at \href{https://huggingface.co/datasets/kpurkayastha/OPRB}{Hugging Face} and \href{https://github.com/kunalpurkayastha/DocRevive}{Github} respectively.
    
    \end{abstract}
\normalfont

\section{Introduction}
Preserving documents is essential for maintaining our cultural, historical, and academic heritage. Over the centuries, countless documents have provided a window into past civilizations, chronicled historical events, and recorded scholarly achievements. Unfortunately, many of these documents suffer from degradation due to environmental exposure, aging materials, and improper handling. Such degradation often turns out as faded ink, physical damage, occluded or distorted text, and in severe cases, complete loss of text in certain regions \cite{zhou2023review}. The restoration of these documents is, therefore, of utmost importance to maintain the integrity and authenticity of the original records. Traditional restoration methods \cite{cannon2003selecting,natarajan1999robust} have primarily focused on enhancing legibility through OCR or manually correcting damaged portions of the text. While these approaches have yielded some success, they typically address only isolated aspects of the degradation problem. For instance, lightweight OCR systems, such as \cite{du2025context,xu2024ote,zhang2024choose}, have demonstrated impressive performance in recognizing text from degraded images. However, they fall short when it comes to inferring missing content or preserving the stylistic elements of the original document due to a lack of contextual awareness and style-matching capability.
\begin{figure}[t]
    \centering
    \includegraphics[width=\columnwidth]{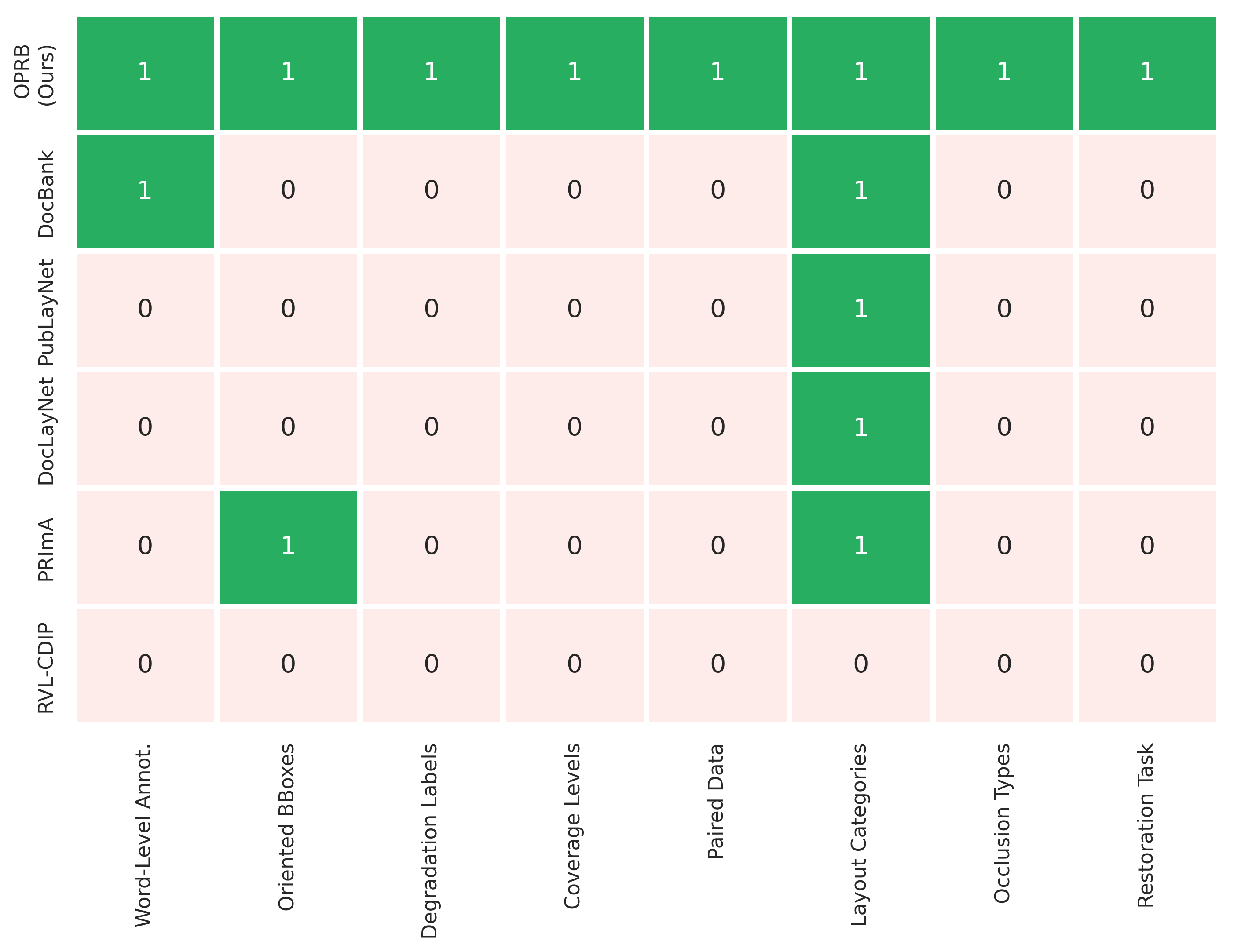}
    \caption{\textbf{Feature completeness across document datasets:} OPRB is the only benchmark that jointly provides degradation-aware labels, paired clean/degraded pages, word-level supervision, and restoration-oriented structure, making it more suitable for document restoration than standard layout-only benchmarks.}
    \label{fig:feature_completeness}
    \vspace{-4mm}
\end{figure}

\noindent Figure~\ref{fig:feature_completeness} explains the need for a new benchmark for restoration rather than only for layout analysis. Existing datasets are useful for tasks such as page parsing or document classification. However, they usually do not provide degraded pages, paired clean targets, explicit occlusion labels, or controlled levels of missing content. In this paper, we introduce a new benchmark dataset, namely the Occluded Pages Restoration Benchmark (OPRB). This is designed around the restoration problem itself. It combines page structure, word-level boxes, and realistic degradation settings in one benchmark, which makes it much more suitable for testing whether a method can truly recover hidden text instead of only detecting or classifying document regions.

\noindent In response to these challenges, our work proposes a novel, unified restoration pipeline that addresses both the semantic and visual aspects of document restoration. The proposed approach begins with an input document that may show various forms of degradation. Such documents might include scanned historical texts, or any printed material that has suffered from prolonged exposure to adverse conditions. Once the degraded document is served as an input in our method, the first step involves processing the image using an OCR module \cite{du2020pp}. This module detects and recognizes the existing text. This initial detection phase is crucial as it sets the stage for the subsequent and more steps in the further restoration process. Following the OCR results, the document is processed by a trained occlusion YOLOv9c \cite{wang2024yolov9} detector, which localizes occlusion patches and classifies them into six degradation types. By intersecting the detected patch bounding boxes with the OCR word layout, the pipeline identifies the precise inter-word gaps that have been obscured, facilitating a more targeted restoration.

Once the degraded regions are accurately identified, the next phase of the pipeline uses a masked language model for contextual text prediction. Unlike traditional methods that rely on simple heuristics \cite{bjerring2022mending} or dictionaries \cite{mittal2020new}, our approach fine-tunes RoBERTa-large \cite{liu2019roberta}, a masked language model, on a document-domain data derived from DocBank \cite{li2020docbank} to infer missing words or characters. The context given to the model is the text before and after the missing part, formatted as a length-conditioned fill-mask prompt. By analyzing the surrounding document text, the model predicts missing gaps with semantically accurate and contextually appropriate completions being computationally efficient. This approach ensures that the restored text aligns with the document’s overall narrative and maintains the intended meaning, even in cases where substantial portions of the text are missing. However, the accurate prediction of text is only one aspect of document restoration. Equally important is the need to preserve the original visual aesthetics of the document. To this end, the pipeline incorporates a text editing model, which is responsible for ensuring that the reconstructed text seamlessly blends with the original document. This module adjusts the key attributes such as font, size, alignment, and spacing, effectively mimicking the original layout and typography. The style transfer process is critical in producing a final output where the restored sections are indistinguishable from the original, thereby maintaining the document’s overall integrity.

We also propose the Unified Context Similarity Metric (UCSM), a metric that evaluates restored text by integrating edit, semantic, and length similarities with a context-based error measure, ensuring an accurate evaluation of the predicted text against the ground truth. Conventional metrics such as the Normalized Edit  (NED) \cite{yujian2007normalized} and Bilingual Evaluation Understudy (BLEU) \cite{papineni2002bleu} neglect contextual cues, making them inadequate for the evaluation of missing text prediction. UCSM was developed to robustly address these nuances and better quantify the restoration, and its full formulation is deferred to the supplementary material.

The final stage of the restoration process is the postprocessing of the retrieved document, where the entire document is refined and formatted for consistency. During this phase, any residual discrepancies between the restored and original content are addressed, ensuring that the final document is both visually cohesive and semantically accurate. This automated, multi-stage workflow minimizes the need for manual intervention, making it a scalable solution for restoring large volumes of degraded documents.

In summary, the key contributions of our work are:
\begin{itemize}
    \item Introduction of a novel synthetic benchmark dataset of 30{,}078 degraded document images, providing a large-scale resource for restoration evaluation where no comparable public benchmark currently exists for occluded document page recovery.
    \item A unified architecture that integrates OCR, occlusion detection, contextual occluded blank prediction, and style-preserving text editing for degraded document restoration. To the best of our knowledge, we are the first one to introduce an unifiied pipeline for document image restoration with explicit missing-text recovery.
    \item A novel metric \emph{UCSM} to evaluate the restoration of the occluded document by jointly considering contextual, semantic, and edit-based information, making it more informative than distance-based metrics for this task.
\end{itemize}

\section{Literature Review}
\myparagraph{Document Restoration:} It has traditionally relied on binarization and denoising techniques like Otsu’s thresholding \cite{Otsu1979} and adaptive binarization \cite{Su2013}, but these methods struggle with complex degradations. On the other hand, with the advent of deep learning approaches \cite{pilligua2024layereddoc,souibgui2022docentr}, the quality of document restoration is significantly improved. Earlier learning-based enhancement models, such as De-GAN \cite{Souibgui2020}, demonstrated the value of generative priors for degraded document cleanup, and DocDiff \cite{Yang2023} later showed that diffusion residual modeling can further improve restoration under complex degradations. Zhang et al. \cite{Zhang2024DocRes} introduced DocRes, a unified model that uses dynamic visual prompting for tasks such as de-warping and de-blurring. In contrast, Yu et al. \cite{Yu2024DocReal} proposed DocReal, an attention-enhanced network for smartphone-captured documents. Generative models have also shown promising development, like Text-DIAE \cite{Souibgui2023TextDIAE}, which is a self-supervised autoencoder for degradation-invariant restoration. Also, a color-aware background model \cite{Zhang2023Shadow} was introduced for enhancing text under shadows. These advancements mark a shift from rule-based to adaptive deep learning methods.

\myparagraph{Optical Character Recognition (OCR):}
OCR has evolved from hand-crafted features \cite{Mori1992} to deep learning, with CRNN \cite{Shi2016} marking a breakthrough. Transformer-based models now set new benchmarks \cite{Li2023TrOCR} introduced TrOCR, combining a Vision Transformer (ViT) encoder with an autoregressive text decoder. Scene text recognition has also advanced with self-supervised learning. Guan et al. \cite{Guan2023SIGA} proposed SIGA, which enhances accuracy by learning glyph structures, while Zhang et al. \cite{Zhang2023LPV} incorporated linguistic priors into vision models. On the other hand, Tang et al. \cite{Tang2023Unified} further used vision, text, and layout processing for structured documents. These advancements shift OCR from text recognition to contextual and structural understanding.

\myparagraph{Document Text Restoration} Despite advances in OCR, text restoration remains challenging, especially for historical documents. Early methods relied on heuristics \cite{Fetaya2020}, while modern approaches use LLMs for post-OCR correction. Thomas et al, \cite{Thomas2024LLMPostOCR} fine-tuned LLMs for error correction and used synthetic OCR errors \cite{Guan2024PostOCR} to train ByT5 for multilingual correction. Similarly, Ensemble methods \cite{Francois2023PostOCR} merged multiple OCR outputs to reduce errors while developing a neural network for restoring ancient Greek inscriptions \cite{Assael2022}. These advancements enhance OCR post-processing and historical text reconstruction.

\myparagraph{Diffusion-Based Text Generation:} Diffusion models  \cite{banerjee2026craftsvg,banerjee2025talediffusion,banerjee2025craftgraffiti} have recently emerged as a powerful method for text image generation and inpainting. TextDiffuser \cite{Chen2023TextDiffuser} proposed a two-stage diffusion framework that first predicts text layouts and then refines the image with a diffusion model. This approach improves the quality and alignment of generated text. More recently, TextCtrl \cite{zeng2025textctrl} introduced prior-guided diffusion for controllable scene text editing, reinforcing the importance of explicit textual control when content and appearance must be preserved simultaneously. Similarly, Zhu et al. \cite{zhu2024text} introduced a global structure-guided diffusion model (GSDM) for inpainting occluded text in images. By using structural priors, the model maintains the correct character structures and arrangements when filling in missing text. Last but not the least, TypeR \cite{Shimoda2024TypeR} proposed a hybrid OCR-diffusion model to detect and correct typos in AI-generated text images. Their approach first detects typographic errors using an OCR model, and then it guides a diffusion-based inpainting method to correct the mistakes while preserving the surrounding visual content. 

\section{Dataset Description}

\begin{figure}[t]
    \centering
    \includegraphics[width=\columnwidth]{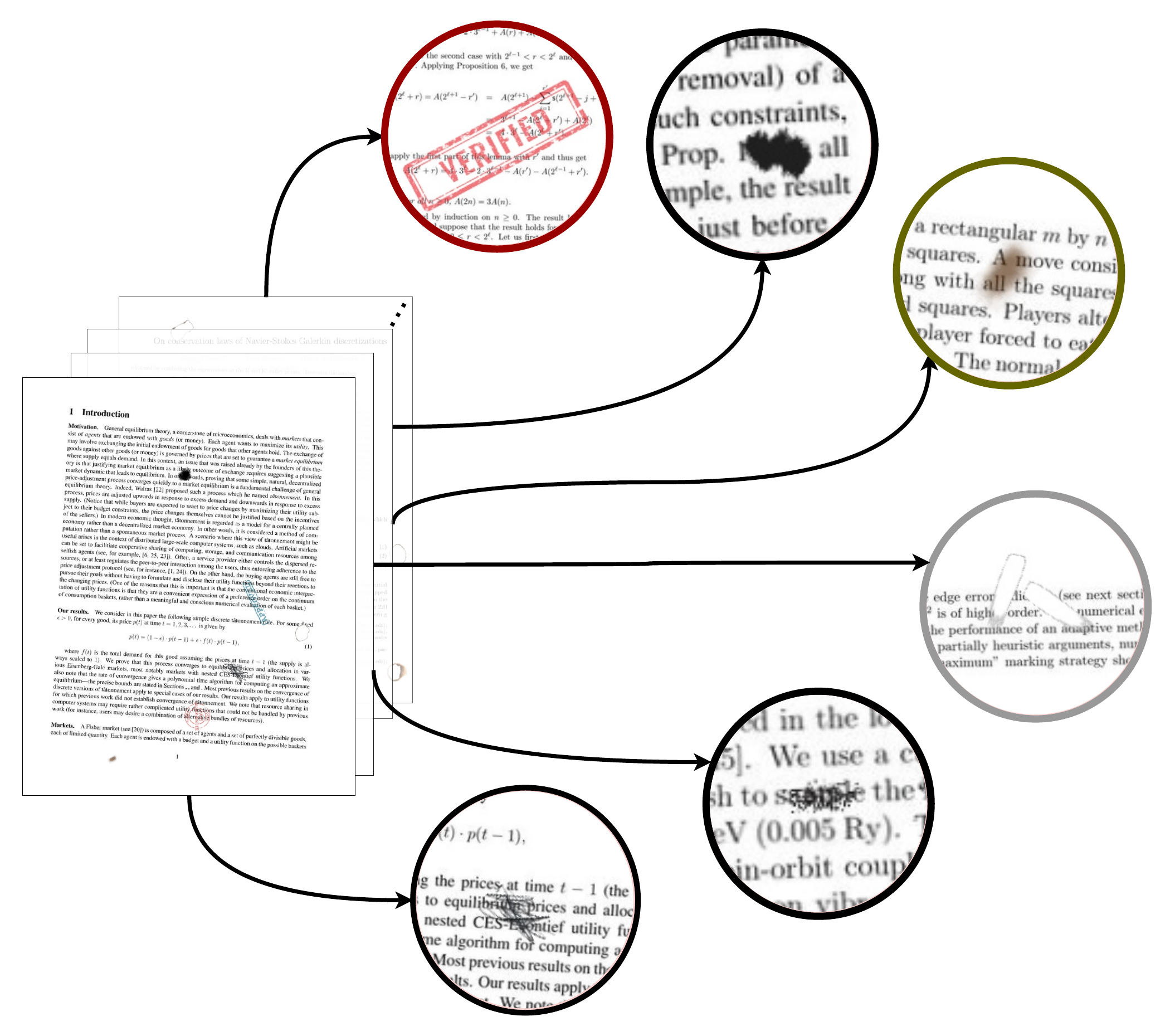}
    \caption{Sample visualization of occluding patches over a document page. Magnified views highlight representative degradations such as stamps, scribbles, black ink, and other occlusions that partially or fully obscure the text.}
    \label{fig:dataset}
    \vspace{-4mm}
 \end{figure}
 
 We introduce a novel benchmark dataset called Occluded Pages Restoration Benchmark(OPRB) designed to evaluate document restoration under a diverse variety of real-world occlusion scenarios. The benchmark contains 30{,}078 degraded document images in total, including 23{,}212 pages with only text and 6{,}866 pages that additionally contain figures or diagrams. OPRB consists of degradation into six occlusion classes spanning both opaque and semi-transparent occlusion modes, which makes it suitable for evaluating restoration under realistic missing-text conditions. To the best of our knowledge, no publicly available benchmark currently offers comparable scale, occlusion diversity, and restoration oriented word-level precision for occluded document page recovery, making OPRB a particularly important resource for fair evaluation in this problem setting.
 
 The benchmark includes four standard classes with controlled coverage levels (Black Ink, Burnt, Whitener, Dust) and two simulation classes (Scribble and Stamp) that models overlapping and semi-transparent overlays. This design is important because document restoration systems must handle not only fully hidden text, but also partially visible content, mixed degradations, and complex page layouts containing non-text elements. Detailed patch generation, annotation refinement, and geometric transformation used to generate the benchmark dataset are deferred to Supplementary Section.~\ref{supp:dataset-details}, where we provide the full construction procedure and algorithms.

 \begin{figure}[t]
    \centering
    \includegraphics[width=\columnwidth]{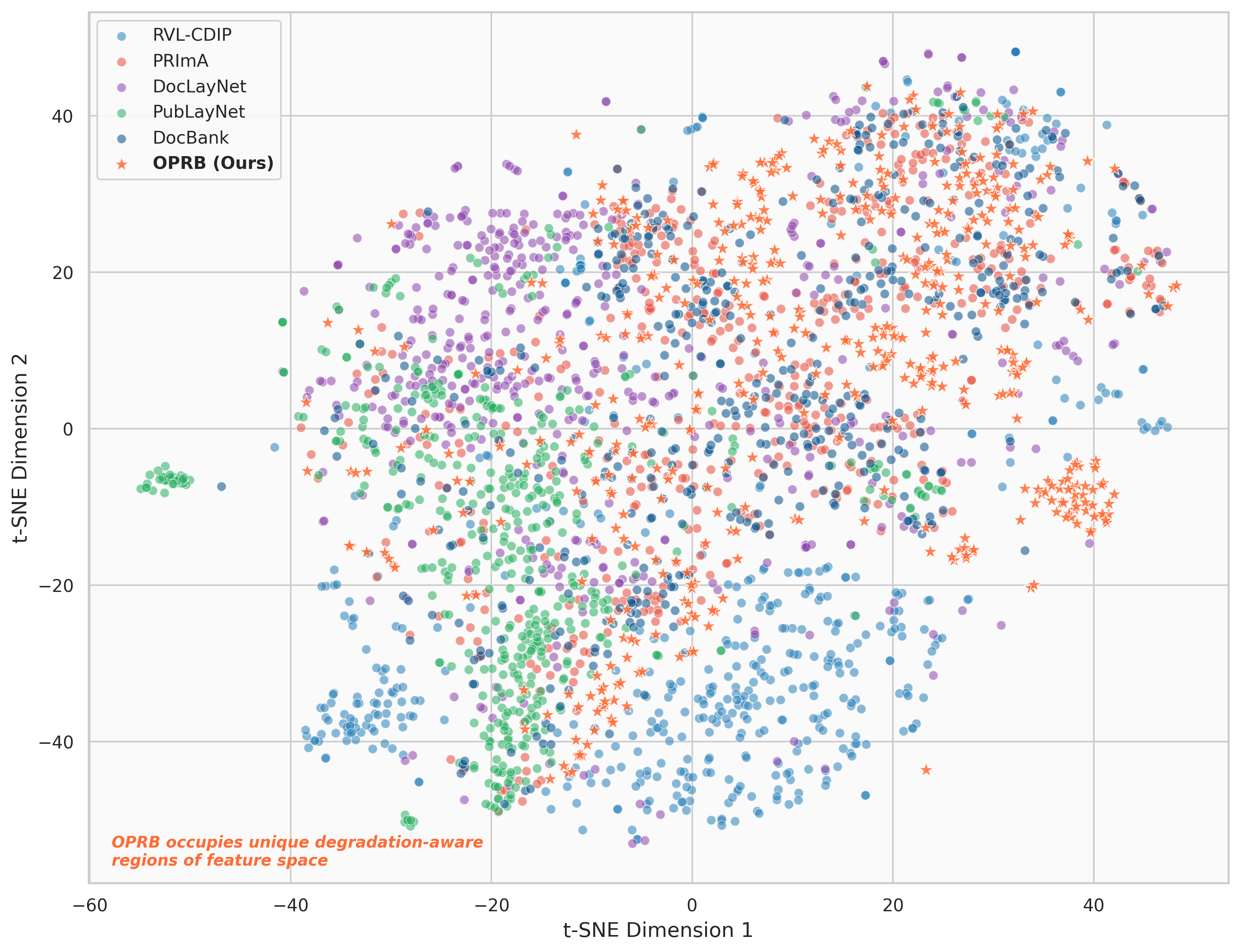}
    \caption{\textbf{t-SNE comparison across document datasets:} OPRB occupies both distinct and shared regions in the shared feature space, indicating that it captures degradation patterns and restoration-relevant document conditions not represented by standard clean-layout benchmarks.}
    \label{fig:tsne_dataset_comparison}
\end{figure}


Figure~\ref{fig:tsne_dataset_comparison} compares OPRB with other document datasets in the t-SNE map. OPRB forms a distinct yet partially overlapping distribution, which is expected because it contains controlled occlusions, paired restoration targets, and degradation specific variation unlike standard clean-layout document page benchmarks. This is desirable for restoration research, since a restoration benchmark should capture realistic visual disruption rather than appear as a lightly perturbed layout dataset. Because OPRB is built from digital research pages, it still shares some feature space with datasets such as DocBank, PubLayNet, and DocLayNet. However, RVL-CDIP forms a more separate cluster, reflecting its more scanned and real-looking document characteristics and has marginal overlapping with the OPRB.

\begin{figure}[t]
    \centering
    \includegraphics[width=\columnwidth]{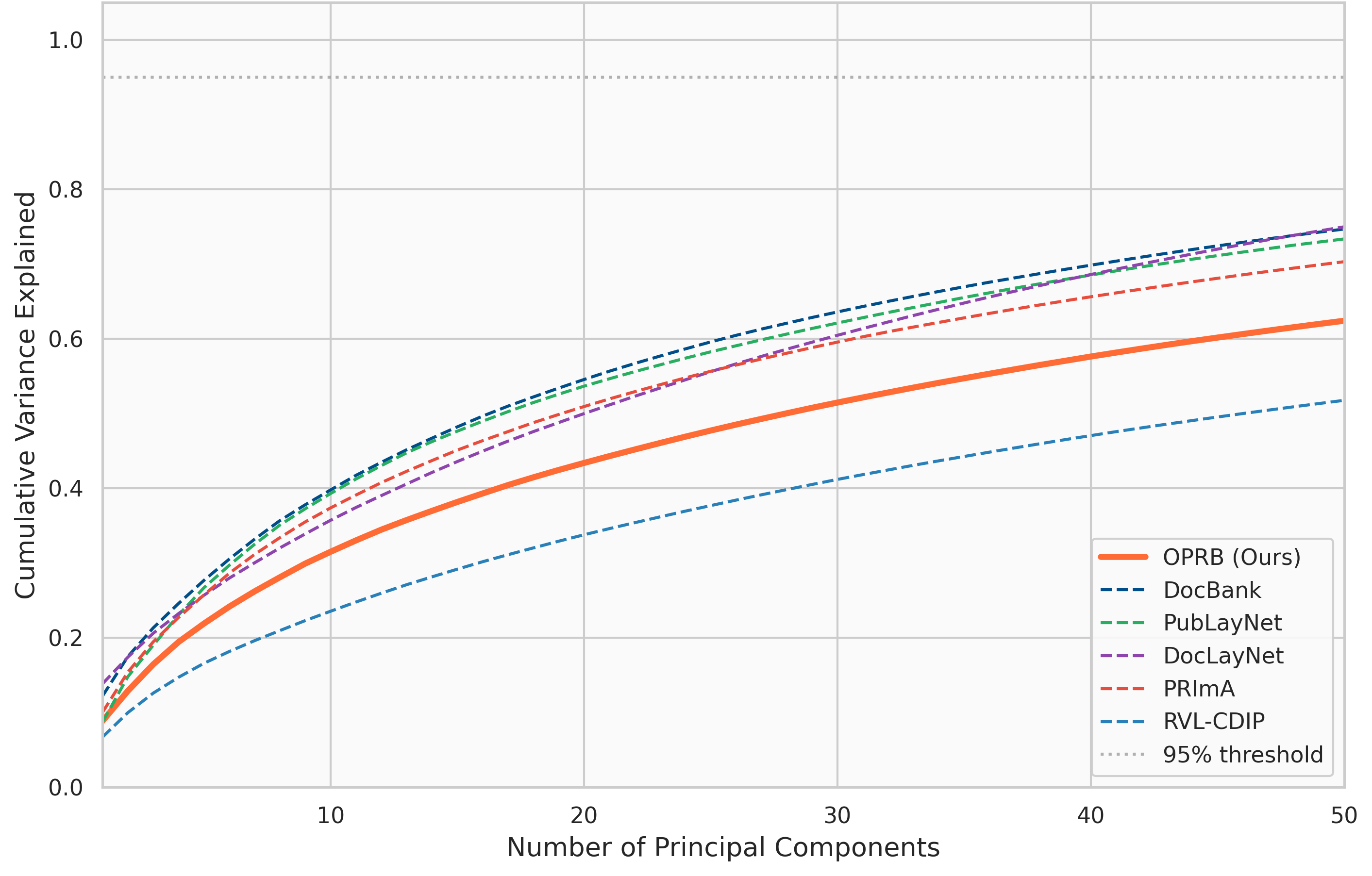}
    \caption{\textbf{PCA variance across datasets:} OPRB exhibits broad variation in its document appearance and degradation patterns, which is important for evaluating restoration methods under diverse conditions.}
    \label{fig:pca_variance}
    \vspace{-4mm}
\end{figure}

The PCA variance plot in Figure~\ref{fig:pca_variance} shows how much of each dataset’s visual diversity is retained as more principal components are added. A slower-rising curve indicates that the dataset contains broader and more complex variation. In our case, OPRB requires more components to capture the same level of variance, suggesting that it covers a richer set of restoration-relevant conditions than standard document benchmarks. However even though the benchmark was formed from the base images of DocBank dataset, still we can see that the variance curve is in between the synthetic-like image datasets and real-like image dataset(RVL-CDIP). This is a good representation of shared properties of the dataset in both synthetic and real domain.

 All annotations are finally exported in the oriented polygon format used for text detection evaluation. The full mathematical description of the annotation update step from the base dataset is also given in the appendix Sec .~\ref{supp:dataset-details}.

\section{Methodology}
We propose a unified pipeline \emph{DocRevive} for document restoration that integrates document text detection, recognition, occlusion detection, blank region extraction, missing text retrieval, and text editing into a unified pipeline. The objective is to restore degraded and occluded document images by first localizing text regions, recognizing the text, identifying gaps (or blanks occured due to occlusion) where text is missing or occluded, and then editing these regions with the target text predicted from context. Given an input document image \(I\), the system predicts text boxes \(B\), recognized strings \(S\), occlusion regions \(P\), restored blanks \(D\), and finally the restored document \(I_{\text{final}}\).


\subsection{Document Text Detection}

Document text detection is performed using FAST \cite{chen2021fast} as it efficiently localizes text regions in the input image \(I\) by generating a set of word-level bounding boxes $(B = \{b_1, b_2, \ldots, b_n\})$, where each \(b_i\) represents the coordinates of a detected word instance. FAST is chosen over other scene text detection models for its robustness in text detection in document images, as the rest of the proposed framework is dependent on accurately localized text regions.

\subsection{Document Text Recognition}

Following detection, the text recognition module uses an autoregressive sequence model \cite{bautista2022scene} to transform each cropped text region into readable strings. For each bounding box \(b_i\), the corresponding text image patch is passed through the recognition network to get a text string \(s_i\). The recognition model is preferred due to its higher prediction accuracy to variations in fonts.

\subsection{Occlusion and Blank Region Extraction} \label{sec:blank-region}

Occlusion patches are first localized using a fine-tuned YOLOv9c detector \cite{wang2024yolov9} trained on the OPRB dataset. Based on the detector ablation in, YOLOv9c is used in the final pipeline because it is the strongest on the opaque degradation classes while remaining highly competitive overall. The detector classifies each patch into one of six degradation types: \emph{Black\_Ink}, \emph{Burnt}, \emph{Whitener}, \emph{Dust}, \emph{Scribble}, and \emph{Stamp}. Opaque classes (\emph{Black\_Ink}, \emph{Burnt}, \emph{Whitener} fully occlude the beneath text, semi-transparent classes (\emph{Dust}, \emph{Stamp}) allow partial visibility of text, and the \emph{Scribble} class simulates scribbling in between text lines.

Once the occluding patches \(P = \{p_1, p_2, \ldots, p_k\}\) are detected, they are intersected with the OCR word layout to identify blank regions that correspond to missing textual gaps. For a text line \(G = \{w_0, w_1, \dots, w_n\}\), each valid blank is represented by the contextual triplet \((\mathrm{PreText}_i, \Box_i, \mathrm{PostText}_i)\), where
\[
\mathrm{PreText}_i \;=\; \mathrm{text}(w_0) \oplus \cdots \oplus \mathrm{text}(w_i), 
\]
\[
\mathrm{PostText}_i \;=\; \mathrm{text}(w_{i+1}) \oplus \cdots \oplus \mathrm{text}(w_n),
\]
and \(\oplus\) denotes string concatenation with spaces. This geometry-aware representation lets the restoration model reason only over blank spaces that are spatially consistent with an occlusion. The detailed blank extraction and token generation procedures are provided in the appendix Algo.~\ref{alg:token_gen}.

\subsection{Missing Token Prediction}

The gap data associated with each emitted blank are fed to a domain-adapted masked language model to predict the missing content. In the current implementation, RoBERTa does not directly consume the multi-blank token string, instead, each blank is converted into a separate masked query derived from its \texttt{pre\_text}, \texttt{post\_text}, and \texttt{max\_chars} fields. We fine-tune RoBERTa-large \cite{liu2019roberta} on a document-specific fill-mask dataset derived from DocBank \cite{li2020docbank}. Pages containing tables, figures, or equations are discarded, and words from textual categories such as paragraph, abstract, title, section, author, caption, and reference are grouped geometrically into text lines. From these lines we construct 900{,}000 training examples and 100{,}000 validation examples. Each example is represented as a length-conditioned masked query:
\[
    \texttt{[K=}N\texttt{] } \mathrm{PreText} \;\texttt{<mask1>}\; \mathrm{PostText}
\]
where $\texttt{[K=}N\texttt{]}$ is a prefix indicating the character length of the masked span, and \texttt{<mask1>} marks the blank position. For multi-blank lines, numbered mask tokens such as \texttt{<mask1>}, \texttt{<mask2>}, and so on are assigned from left to right and aligned with the corresponding ground-truth labels. During fine-tuning, these numbered mask tokens are mapped to RoBERTa's standard \texttt{<mask>} token and the masked language modelling loss is applied only at the blank positions.

At inference time, this fine-tuned model remains fully offline and document domain specific. For each blank index $\alpha$, the model receives
\[
    {P}_\alpha = \texttt{[K=}M_\alpha\texttt{] } \mathrm{PreText}_\alpha \;\texttt{<mask>}\; \mathrm{PostText}_\alpha
\]
where $M_\alpha$ is the blank-specific \texttt{max\_chars} estimate. The model returns the top-1 fill for the blank. For multi-word fills, decoding is performed iteratively for approximately $\max(1,\mathrm{round}(M_\alpha / 6))$ steps, feeding the partial prediction back into the left context at each step. The resulting predicted text $D_\alpha$ is then used as the target text input for the subsequent text editing module.

\subsection{Document Text Editing}

In the document text editing stage we use a diffusion-based text editing model \cite{zeng2025textctrl} built on a UNet–VAE–ControlNet backbone with DDIM sampling. A style patch \(I_s\) is extracted from the surrounding context of each blank region, and the target text \(D_{t}\) predicted by the language model is used to guide the text restoration. The editing model introduces a glyph-aware attention mechanism that conditions the diffusion process on both the visual style of \(I_s\) and the character-level glyph information of \(D_t\) via an ABINet-based vision monitor. The model takes \(I_s\) and \(D_{t}\) as inputs to generate an edited patch \(I_e\) that maintains the visual style of the document while integrating the desired target text in the same style as the rest of the document. This stage ensures that the edited regions are consistent in terms of font, color, and background with the rest of the document, preserving the overall integrity of the document image.

\subsection{Unified Document Restoration}

\begin{figure*}[htbp]
  \centering
  \includegraphics[width=\textwidth]{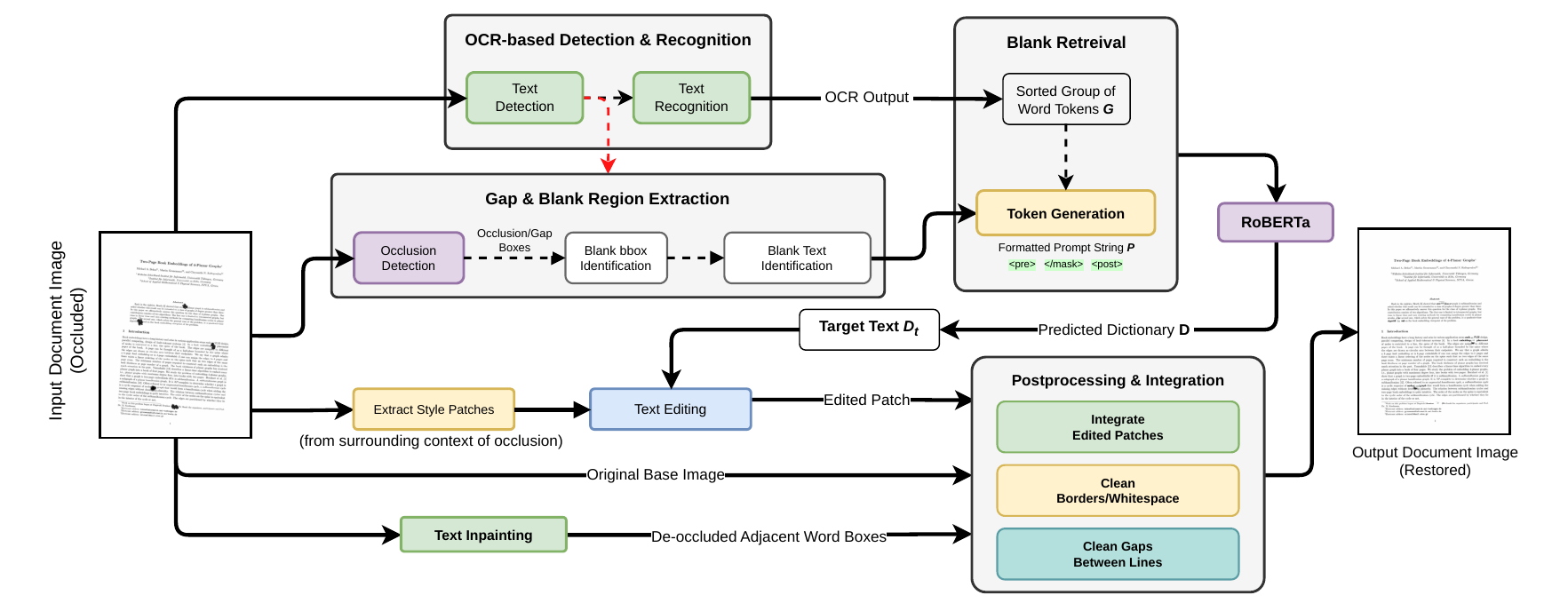}
  \caption{\textbf{DocRevive:} a unified model architecture for the proposed document restoration pipeline. Given a degraded document image, the framework first performs OCR, followed by occlusion-aware blank region extraction to localize missing text areas. The surrounding word context is then grouped and converted into formatted prompt tokens, which are processed by the RoBERTa masked language model to predict the missing content. In parallel, style patches from nearby regions guide the text-editing branch, while text inpainting helps recover occluded neighboring text regions. Finally, postprocessing integrates the edited patches, cleans the gaps between lines, and generates a restored document image.}
  \label{fig:dataset_samples}
\end{figure*}

Our unified document restoration pipeline is designed to reconstruct degraded or occluded document images by first extracting occluded regions and then restoring them using a diffusion-based text editing model. The overall pipeline, illustrated in Figure~\ref{fig:dataset_samples}, can be summarized as
\[
I \xrightarrow{f_{\text{det}}} B \xrightarrow{f_{\text{rec}}} S \xrightarrow{f_{\text{occ}}} P \xrightarrow{f_{\text{gap}}} \mathcal{G} \xrightarrow{f_{\text{mask}}} D \xrightarrow{f_{\text{edit}}} I_{\text{final}},
\]
where \(f_{\text{det}}\) and \(f_{\text{rec}}\) denote FAST and PARSEQ, \(f_{\text{occ}}\) denotes the YOLOv9c occlusion detector, \(f_{\text{gap}}\) extracts blank regions from OCR-occlusion intersections, \(f_{\text{mask}}\) predicts the missing text, and \(f_{\text{edit}}\) denotes text editing and compositing the final restored document.


After the missing blanks are predicted, the text editor renders the target text in a style patch extracted from the original page. The edited patch \(I_e\) is conditioned on both the target text and the surrounding visual appearance so that the restored region remains style-consistent with the original document.

To ensure consistency between the restored document and the original, a postprocessing step is performed after text editing. A text inpainting module \cite{zhu2024text} is applied on the adjacent word bounding boxes of the detected blank to de-occlude visible remains or possible blobs, after which the edited patches are cleaned and integrated back into the page. Detailed postprocessing and the full geometry-driven restoration procedure are provided in the appendix Sec.~\ref{supp:method-details}.


\section{Experimental Evaluation}
In this section, we introduce our own metric to evaluate the unified document restoration pipeline. We also evaluate both the individual modules of our document restoration pipeline and the overall system. The evaluation is divided into two subsections. In the first part, \emph{Phasewise Evaluation}, we assess the performance of each module used in our unified approach individually. In the second part, \emph{Unified Evaluation}, we measure the overall restoration quality of the integrated system using metrics that reflect both visual fidelity and semantic accuracy.

\subsection{UCSM Evaluation Metric}

We propose a Unified Context Similarity Metric (UCSM) to evaluate the similarity between the predicted string \(P\) for the blank space and the ground truth string \(GT\). In our problem, a useful metric must go beyond raw string overlap. Basically, a prediction can have a modest edit distance and still be semantically wrong, while another prediction can be lexically imperfect but contextually reliable. UCSM is designed for this specific missing-text restoration scenario by combining lexical similarity, semantic similarity, and length consistency, while also accounting for how predictable the ground-truth answer is from its surrounding context.

This contextual component is necessary because not all blanks are equally difficult. If the surrounding text makes the missing span almost obvious, incorrect predictions should be penalized more strongly and if the context is genuinely ambiguous or hard to predict the blank space, the metric should be more forgiving than pure distance-based measures such as NED or BLEU. In this way, UCSM better reflects the actual restoration objective than conventional metrics for our task. The final score is defined as
\begin{equation}
UCSM = S(P,GT)^{(1 - E_{\text{context}})}.
\label{eq:UCSM}
\end{equation}

Here, \(S(P,GT)\) is the geometric mean of three different similarity terms, \(P\) is the predicted missing blank text, \(GT\) is the ground-truth text. $E_{\text{context}}$ is the predictibility of the blank region. A lower value of $E_{\text{context}}$ indicates that the missing span is highly predictable from context, whereas a value near $1$ indicates low predictability. The full mathematical formulation, calibration terms, and detailed derivation are provided in Supplementary Sec.~\ref{supp:ucsm-details}.

\subsection{Qualitative Analysis}
For the unified pipeline, we evaluate overall restoration quality using SSIM for visual fidelity, edit distance (ED) for textual difference, and document-level UCSM for semantic consistency. Table~\ref{tab:e2e} summarizes the performance of the proposed framework on the OPRB dataset.
\begin{figure*}[htbp]
  \centering
  \includegraphics[width=\textwidth]{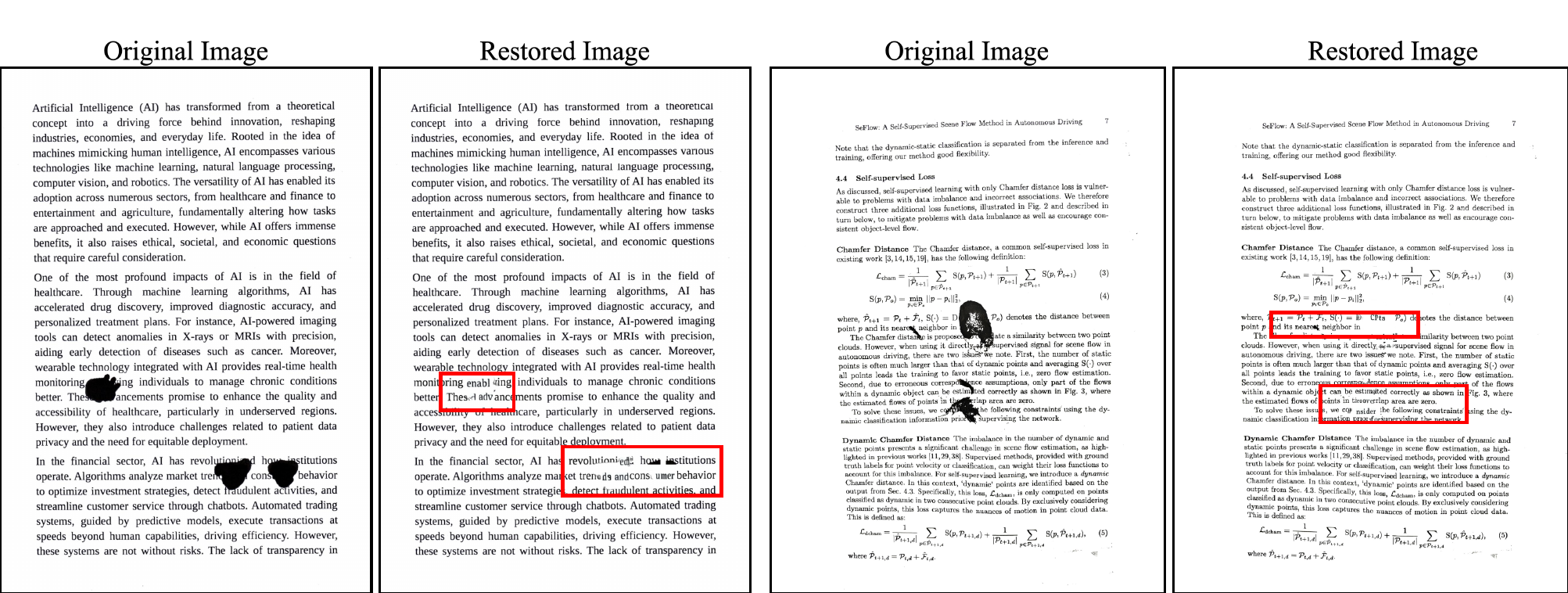}
  \caption{Visualization of the proposed framework on a real scanned document degraded using whiteboard marker ink. The document image in \textbf{Left:} was destroyed using concentrated ink, whereas the document in \textbf{Right:} was destroyed using diluted ink.}
  \label{fig:viz}
  \vspace{-4mm}
\end{figure*}

\subsection{Ablation Studies}


We perform a class-wise occlusion detector ablation training for all detectors on the six degradation classes. Table~\ref{tab:detector_ablation_nomixed} reports the per-class mAP50 scores for Black Ink, Burnt, Whitener, Dust, Scribble, and Stamp, together with the average across classes. with YOLO26n performing substantially worse, YOLO11n achieves the highest average mAP50, while YOLOv8s is the second-best model overall. However, YOLOv9c gives the strongest results on Black Ink and Whitener and ties for the top score on Burnt and Stamp. Since downstream restoration is driven primarily by accurate localization of opaque occlusions that fully remove text, we adopt YOLOv9c as the occlusion detector in the final DocRevive pipeline. These results indicate that newer YOLO design choices significantly improve robustness for fine-grained occlusion localization.

\begin{table}[htbp]
\centering
\small
\caption{Class-wise occlusion detector ablation after removing the Mixed class. Values report mAP50 (\%). \textbf{Bold} texts represent best results whereas \underline{underlined} texts represent second best results.}
\vspace{-2mm}
\label{tab:detector_ablation_nomixed}
\resizebox{\columnwidth}{!}{
\begin{tabular}{lccccccc}
\hline
\textbf{Method} & \multicolumn{3}{c}{\textbf{Opaque Classes}} & \multicolumn{2}{c}{\textbf{Transparent Classes}} & \textbf{Special} & \textbf{Average} \\ \cline{2-8}
 & \textbf{B.Ink} & \textbf{Burnt} & \textbf{Whitener} & \textbf{Dust} & \textbf{Stamp} & \textbf{Scribble} & \textbf{Avg.} \\
\hline
YOLO26n \cite{sapkota2509yolo26} & 93.701 & 92.254 & 90.880 & 71.175 & 86.802 & 86.103 & 86.819 \\
YOLOv8n \cite{yaseen2025yolov8} & 99.452 & \underline{99.498} & 99.483 & 90.498 & \textbf{99.500} & 99.270 & 97.950 \\
YOLOv8s \cite{yaseen2025yolov8} & \underline{99.493} & \textbf{99.500} & \underline{99.496} & 91.895 & \underline{99.487} & \textbf{99.442} & \underline{98.219} \\
\textbf{YOLOv9c} \cite{wang2024yolov9} & \textbf{99.500} & \textbf{99.500} & \textbf{99.497} & 90.675 & \textbf{99.500} & \underline{99.431} & 98.017 \\
YOLOv10b \cite{wang2024yolov10} & 99.490 & 99.487 & 99.454 & \underline{92.153} & 99.079 & 99.288 & 98.159 \\
\underline{YOLO11n} \cite{khanam2024yolov11} & \underline{99.493} & 99.493 & 99.482 & \textbf{92.157} & \textbf{99.500} & 99.367 & \textbf{98.249} \\
\hline
\end{tabular}
}
\end{table}

\noindent To compare the text restoration quality used in our framework, we evaluate a fine-tuned RoBERTa model and a fine-tuned Qwen3-0.6B causal language model on a 6000-sample test subset drawn randomly. We report UCSM, mean edit distance (ED), character error rate (CER), and exact-match rate. Both models are fine-tuned on 1M samples of pre, mask, and post texts from our OPRB benchmark. Overall, RoBERTa achieves a mean UCSM of 0.6102 across all six buckets, consistently outperforming Qwen3-0.6B on this metric, while Qwen3-0.6B demonstrates a lower CER across all intervals, indicating complementary strengths between the two approaches. Table~\ref{tab:lm_ablation_overall} summarizes the overall comparison, while Table~\ref{tab:lm_ablation_buckets} shows the bucket-wise trend over consecutive 1000-sample intervals.

\begin{table}[htbp]
\centering
\small
\caption{RoBERTa \cite{liu2019roberta} vs. Qwen3 \cite{qwen3_2025} text prediction ablation on a 6000-sample validation subset. \textbf{Bold} represent best results whereas \underline{underlined} represent second best results. Higher UCSM and exact-match are better, while lower ED and CER are better.}
\vspace{-2mm}
\label{tab:lm_ablation_overall}
\resizebox{\linewidth}{!}{%
\begin{tabular}{lcccc}
\hline
\textbf{Model} & \textbf{UCSM $\uparrow$} & \textbf{ED $\downarrow$} & \textbf{CER $\downarrow$} & \textbf{Exact Match (\%) $\uparrow$} \\
\hline
\textbf{RoBERTa-large (fill-mask)} \cite{liu2019roberta} & \textbf{0.6102} & \underline{8.2662} & \underline{1.3762} & \textbf{21.60} \\
Qwen3-0.6B (causal LM) \cite{qwen3_2025} & \underline{0.1240} & \textbf{7.0928} & \textbf{1.0807} & \underline{0.15} \\
\hline
\end{tabular}
}
\end{table}

\begin{table}[htbp]
\centering
\small
\caption{Bucket-wise comparison between fine-tuned RoBERTa \cite{liu2019roberta} and Qwen3 \cite{qwen3_2025} on a validation fold. Each bucket contains 1000 consecutive expanded validation samples. \textbf{Bold} represent best results whereas \underline{underlined} represent second best results.}
\vspace{-2mm}
\label{tab:lm_ablation_buckets}
\resizebox{\columnwidth}{!}{
\begin{tabular}{lcccccc}
\hline
\textbf{Sample Set} & \multicolumn{2}{c}{\textbf{UCSM $\uparrow$}} & \multicolumn{2}{c}{\textbf{Exact Match (\%) $\uparrow$}} & \multicolumn{2}{c}{\textbf{ED $\downarrow$}} \\ \cline{2-7}
 & \textbf{RoBERTa} & \textbf{Qwen3} & \textbf{RoBERTa} & \textbf{Qwen3} & \textbf{RoBERTa} & \textbf{Qwen3} \\
\hline
0-1000    & \textbf{0.6131} & \underline{0.1221} & \textbf{22.30} & \underline{0.10} & \underline{8.4530} & \textbf{7.3240} \\
1001-2000 & \textbf{0.6201} & \underline{0.1243} & \textbf{22.40} & \underline{0.10} & \underline{8.2380} & \textbf{7.0400} \\
2001-3000 & \textbf{0.6093} & \underline{0.1154} & \textbf{21.50} & \underline{0.40} & \underline{8.2000} & \textbf{7.0470} \\
3001-4000 & \textbf{0.6175} & \underline{0.1335} & \textbf{23.00} & \underline{0.20} & \underline{8.2210} & \textbf{7.0370} \\
4001-5000 & \textbf{0.6046} & \underline{0.1317} & \textbf{19.90} & \underline{0.10} & \underline{8.3670} & \textbf{7.0380} \\
5001-6000 & \textbf{0.5966} & \underline{0.1173} & \textbf{20.50} & \underline{0.00} & \underline{8.1180} & \textbf{7.0710} \\
\hline
\end{tabular}}
\end{table}

RoBERTa consistently outperforms Qwen3 in UCSM across all six buckets, remaining stable in the 0.54-0.56 range, whereas Qwen3 stays near 0.12-0.13. The exact-match rate also favors RoBERTa, achieving 21.6\% overall compared with only 0.15\% for Qwen3. Although Qwen3 shows a slightly lower mean raw edit distance, these predictions show that it often produces short but semantically incorrect continuations where these outputs can keep the absolute edit distance modest while still giving very poor normalized similarity. In contrast, RoBERTa occasionally makes longer errors, but its predictions are substantially more reliable to the target text overall, which is reflected by its much higher UCSM and exact-match rate. This ablation therefore supports our choice of the RoBERTa fill-mask model as the more reliable text restoration backbone in the current pipeline.

\subsection{Phasewise Evaluation}

The performance of the text detection module is evaluated by calculating precision, recall, and F1 scores. Table~\ref{tab:detection} provides the results for different detection methods evaluated on the OPRB dataset. The recall is slightly higher than the precision due to the detection of false bounding boxes because of the document noise or fragments of scribbles, stamps(with text in it) and other occlusion patches. Similarly, text recognition performance is measured in terms of accuracy, as shown in Table~\ref{tab:recognition}. The comparison study of the text detection experiment was performed for a total of 5520 degraded document samples, while the recognition experiment was performed for 54614 cropped text samples.


\begin{table}[htbp]
\centering
\caption{Text detection performance on OPRB dataset. \textbf{Bold} texts represent best results whereas \underline{underlined} texts represent second best results. Here DB means Differentiable Binarization}
\vspace{-2mm}
\label{tab:detection}
\resizebox{\columnwidth}{!}{%
\begin{tabular}{lcccc}
\hline
\textbf{Method and Variant} & \textbf{Precision} & \textbf{Recall} & \textbf{F1-Score} & \textbf{mIoU} \\
\hline
LinkNet-ResNet18 \cite{chaurasia2017linknet}         & 89.46 & 92.52 & 90.97 & 74.00 \\
LinkNet-ResNet34 \cite{chaurasia2017linknet}         & 87.38 & 91.07 & 89.18 & 74.00 \\
LinkNet-ResNet50 \cite{chaurasia2017linknet}         & 88.58 & 93.56 & 91.00 & 74.00 \\
DB-ResNet50 \cite{liao2020real}              & 89.33 & \textbf{95.27} & 92.20 & 76.00 \\
DB-MobileNetV3L \cite{liao2020real}   & 88.60 & 93.80 & 91.13 & 74.00 \\
FAST-tiny \cite{chen2021fast}                 & 90.68 & 94.63 & 92.61 & 78.00 \\
FAST-small \cite{chen2021fast}                & \underline{92.03} & 94.44 & \underline{93.22} & \underline{79.00} \\
\textbf{FAST-base \cite{chen2021fast} (Our Baseline) }                & \textbf{93.08} & \underline{94.88} & \textbf{93.97} & \textbf{80.00} \\
\hline
\end{tabular}
}
\end{table}


\begin{table}[htbp]
\centering
\small
\caption{Text recognition performance on OPRB dataset. \textbf{Bold} represent best whereas \underline{underlined} represent second best results.}
\vspace{-2mm}
\label{tab:recognition}
\resizebox{\columnwidth}{!}{%
\begin{tabular}{lccc}
\hline
\textbf{Method} & \textbf{Accuracy} & \textbf{1-NED} & \textbf{Confidence} \\
\hline
CLIPOCR \cite{wang2023symmetrical} & \underline{94.55} & \underline{97.71} & \textbf{96.34} \\
DATR \cite{purkayastha2025datr}    & 89.97 & 96.19 & 90.94 \\
\textbf{PARSEQ \cite{bautista2022scene} (Our Baseline)}  & \textbf{95.42} & \textbf{98.03} & \underline{95.85} \\
\hline
\end{tabular}
}
\vspace{4pt}
\end{table}

\subsection{Unified Evaluation}

For the unified pipeline, we evaluate the overall restoration performance by measuring both visual quality and semantic consistency. We compute SSIM to evaluate the structural similarity of the restored document. FID to indicate the distribution similarity, and a document-level UCSM score is used to assess semantic alignment. Table~\ref{tab:e2e} shows the results of the unified method.

\begin{table}[htbp]
\centering
\small
\caption{Performance comparison across different occlusion types on DocRevive. Higher SSIM and UCSM indicate better perceptual quality, while lower FID indicates better distribution similarity.}
\vspace{-2mm}
\label{tab:e2e}
\resizebox{0.7\columnwidth}{!}{%
\begin{tabular}{lccc}
\hline
\textbf{Type} & \textbf{SSIM $\uparrow$} & \textbf{FID $\downarrow$} & \textbf{UCSM $\uparrow$} \\
\hline
Black Ink  & 0.6844 & 19.10 & 0.6667 \\
Burnt      & 0.6859 & 20.50 & \textbf{0.9631} \\
Dust       & 0.7230 & 20.08 & 0.5138 \\
Scribble   & 0.6748 & 19.41 & 0.6352 \\
Stamp      & 0.6329 & 38.11 & 0.3717 \\
Whitener   & \textbf{0.7560} & \textbf{14.36} & 0.5104 \\
\hline
\textbf{Average} & 0.6928 & 15.70 & 0.6102 \\
\hline
\end{tabular}
}
\end{table}

Table~\ref{tab:e2e} shows that the proposed method performs robustly across different occlusion types. Whitener gives the best visual restoration, with the highest SSIM (0.7560) and lowest FID (14.36), while Burnt achieves the strongest semantic recovery with the highest UCSM (0.9631). Black Ink and Scribble give balanced performance overall. However, Stamp is the most challenging case, with the lowest SSIM and UCSM and the highest FID, indicating that stamp-like occlusions remain harder to restore effectively. The Stamp and Dust are still considering some blanks are mostly due to OCR failure and double columned documents.

\section{Conclusion}
This paper presents an occluded document restoration framework that combines OCR, a fine-tuned YOLOv9 occlusion detector, geometry-driven blank extraction, a document-based fine-tuned RoBERTa masked model, and diffusion-based text editing to recover missing content while preserving document style. Along with the proposed UCSM metric, which unifies edit, semantic, and contextual similarity into a single evaluation score, and a synthetic benchmark dataset spanning six occlusion types, the framework employs robust semantic and visual restoration. Experimental results across all occlusions demonstrate that the pipeline achieves competitive scores, confirming that multimodal processing is an effective design choice for this problem. The framework offers a practical step toward reducing effort in restoring degraded documents. In future we will be handling more complex layouts, real-world archival collections, and fully unified restoration pipeline.

\section{Limitation}
The current framework expects single-column layouts with horizontal line images. Therefore, multi-column documents, tables, and mixed text-figure regions may cause blank boundaries to span incorrectly. The fixed context window of RoBERTa does not enforce long-range coherence across multiple nearby blanks. Extending evaluation and developing a more robust method to tackle this problem remains as our future work.

\section{Impact Statement}
Although our work is intended to support document preservation and archival recovery, it may also have negative impacts if used without human oversight. Since the system reconstructs missing text from context, it can generate plausible but incorrect content, which may unintentionally alter the meaning of historical, legal, or administrative documents. In addition, similar restoration methods could be misused to recover intentionally obscured or sensitive information. Therefore, restored outputs should be treated as assistive reconstructions and verified by domain experts.

\myparagraph{Acknowledgements:}
This piece of research was carried out with the support of the following granted projects: SGR Grant 2021 SGR 01559 from the Catalan Government, GRAIL PID2021-126808OB-I00 and from FEDER/UE, SUKIDI PID2024-157778OB-I00 grants from the Spanish Ministry of Science and Innovation, with the support of Cátedra UAB-Cruïlla grant TSI-100929-2023-2 from the Ministry of Economic Affairs and Digital Transformation of the Spanish Government.

{
    \small
    \bibliographystyle{ieeenat_fullname}
    \bibliography{main}
}
\clearpage
\setcounter{page}{1}
\maketitlesupplementary

\section{Dataset Construction Details}
\label{supp:dataset-details}

This supplementary section provides the full construction details of the Occluded Pages Restoration Benchmark (OPRB). In the current generator, we choose $N$ unique source pages per class-level.

We introduce a novel benchmark dataset called Occluded Pages Restoration Benchmark (OPRB) designed to evaluate document restoration under a variety of real-world like simulated occlusions. The dataset consists of degradation into six occlusion classes, each with distinct visual and semantic properties. Four \emph{standard} classes, Black Ink, Burnt, Whitener, Dust, target a quantitatively controlled pixel coverage level $T\%$ of the total document area $A_D$, with the current generator using three target levels $T \in \{0.5, 1.0, 1.5\}$. Opaque classes (Black Ink, Burnt, Whitener) use fully opaque patches, whereas Dust uses semi-transparent patches at opacity $0.65$, through which underlying text remains partially readable. Two additional \emph{simulation} classes, Scribble and Stamp, use the placement logic and do not target a numeric coverage percentage. Source images are partitioned such that no image appears in more than one class-level combination, ensuring all evaluation splits are strictly non-overlapping.

For standard classes, between $n \sim \mathcal{U}\{3, 7\}$ patches are selected and placed sequentially to approach the target covered area $A_T = \frac{T}{100}\,A_D$. Coverage is tracked pixel-precisely via a binary coverage mask $M_\text{cov}$ of the same resolution as the document, and a parallel occlusion mask $M_\text{occ}$ used exclusively for ground-truth refinement, which is populated only for opaque patches. For each patch $P$ with effective non-transparent pixel area $A_P$, the placement scale during the initial pass is computed as
\begin{equation}
    s = \sqrt{\frac{A_R}{A_P}},
\end{equation}
where $A_R = \frac{A_T - A_O}{k_{\text{left}}}$ is the per-patch share of the remaining uncovered area, $k_{\text{left}}$ is the number of initial-pass patches left to place, and $A_O = \lvert M_\text{cov} \rvert$ denotes the cumulative covered pixel count at the time of placement. The scale is clamped to $[0.005, 10.0]$ to prevent degenerate patches. After scaling, each initial-pass patch is rotated by $\theta \sim \mathcal{U}[0^\circ, 360^\circ]$ and placed at a position drawn uniformly from $\mathcal{U}\!\left[-\tfrac{w_P}{4},\, W - \tfrac{3w_P}{4}\right] \times \mathcal{U}\!\left[-\tfrac{h_P}{4},\, H - \tfrac{3h_P}{4}\right]$, permitting partial out-of-frame placement to simulate edge occlusions. If the target is not reached within the initial $n$ patches, up to six top-up passes add further patches scaled to the full residual area $A_T - A_O$ until $A_O \geq 0.95\,A_T$ these top-up patches are placed fully within the document bounds.

The simulation class simulates crossing and scribbling out words by placing handwritten scribble patches over random word bounding boxes. Between 5 and 6 words per image are targeted, sampled uniformly as $n \sim \mathcal{U}\{5, 6\}$, subject to validity filters, the word must belong to the \texttt{paragraph} class of the source ground truth, have length greater than two characters, contain at least one alphabetic character, not consist solely of symbols or digits, have a minimum bounding box width of $15$\,px, and not lie within a figure or equation exclusion zone. Each selected scribble is resized to match the word bounding box with a small random oversize factor in width $\mathcal{U}[1.00, 1.15]$ and height $\mathcal{U}[1.00, 1.20]$, placed with slight positional jitter, and tilted by $\pm 10^\circ$ for a natural handwritten appearance. Scribbled words are entirely removed from the ground truth. Whereas, the Stamp class places exactly one semi-transparent stamp (opacity $0.60$) per document in one of two modes chosen at random (a) \emph{centre} stamp scaled to $\mathcal{U}[20, 35]\%$ of the document height and centred on the page, or (b) \emph{corner} stamp scaled to $\mathcal{U}[8, 12]\%$ of the document height placed at the bottom-left or bottom-right corner with a $3\%$ margin, both tilted by $\pm 15^\circ$. Because stamps are semi-transparent, the ground truth is left entirely unmodified.

Word-level bounding boxes are refined against $M_\text{occ}$ via the column-wise procedure described in Algorithm~\ref{alg:gt_refinement}. For each surviving word, a column occupancy vector $\mathbf{c} \in \mathbb{R}^{w}$ is computed over the bbox region, where $c_j = \frac{1}{h}\sum_{i} M_\text{occ}[y_1{:}y_2,\, x_1+j]$. The valid horizontal extent is found by scanning inward from both sides to locate columns with occupancy below a threshold $\tau_\text{col} = 0.30$, and trimming any interior dense wall where $c_j \geq \tau_\text{dense} = 0.90$. A word is removed if no valid column extent exists, if the global occlusion ratio of the trimmed region exceeds $\tau_\text{global} = 0.50$, or if the trimmed bbox falls below minimum dimensions ($w < 10$\,px, $h < 10$\,px, or area $< 150$\,px$^2$). This refinement applies only to standard classes and the simulatinf scribbles removes the scribbled words entirely, and Stamp and Dust leaves all annotations intact.

\begin{algorithm}[t]
\caption{Column-wise Ground-Truth Refinement (Standard Mode)}
\label{alg:gt_refinement}
\begin{algorithmic}[1]
\State \textbf{Input:} Word bboxes $\{B_i\} = \{(x_1, y_1, x_2, y_2)\}_i$; occlusion mask $M_\text{occ}$
\State \textbf{Constants:} $\tau_\text{col} = 0.30$,\ $\tau_\text{dense} = 0.90$,\ $\tau_\text{global} = 0.50$,\ $w_\text{min} = 10$,\ $h_\text{min} = 10$,\ $A_\text{min} = 150$
\For{each $B_i = (x_1, y_1, x_2, y_2)$}
    \State $\text{sub} \leftarrow M_\text{occ}[y_1{:}y_2,\; x_1{:}x_2]$;\quad $bw \leftarrow x_2 - x_1$;\quad $bh \leftarrow y_2 - y_1$
    \If{$\lvert \text{sub} \rvert / (bw \cdot bh) < 0.05$}
        \State Keep $B_i$ unchanged
        \State \textbf{continue}
    \EndIf
    \State Compute column occupancy $c_j \leftarrow \frac{1}{bh}\sum_i \text{sub}[i, j]$ for $j = 0, \ldots, bw - 1$
    \State $l \leftarrow \min\{j : c_j < \tau_\text{col}\}$;\quad $r \leftarrow \max\{j : c_j < \tau_\text{col}\}$
    \If{no valid $l$ or $r$, or $r < l$}
        \State \textbf{Remove} $B_i$
        \State \textbf{continue}
    \EndIf
    \For{$j = l$ to $r$}
        \If{$c_j \geq \tau_\text{dense}$}
            \If{$j = l$}
                \State \textbf{Remove} $B_i$
                \State \textbf{break}
            \Else
                \State $r \leftarrow j - 1$
                \State \textbf{break}
            \EndIf
        \EndIf
    \EndFor
    \State $fw \leftarrow r - l + 1$
    \If{$\lvert \text{sub}[:, l{:}r{+}1]\rvert / (fw \cdot bh) > \tau_\text{global}$}
        \State \textbf{Remove} $B_i$
        \State \textbf{continue}
    \EndIf
    \If{$fw < w_\text{min}$ or $bh < h_\text{min}$ or $fw \cdot bh < A_\text{min}$}
        \State \textbf{Remove} $B_i$
        \State \textbf{continue}
    \EndIf
    \State Update $B_i \leftarrow (x_1 + l,\; y_1,\; x_1 + r + 1,\; y_2)$
\EndFor
\State \textbf{Output:} Refined bounding boxes $\{B_i\}$
\end{algorithmic}
\end{algorithm}

After occlusion and ground-truth refinement, all images across all classes are subjected to a random global rotation $\varphi \sim \mathcal{U}[-5.0^\circ, +5.0^\circ]$ to simulate realistic scan misalignment, using full-canvas expansion with white fill so that no content is clipped. The rotation matrix
\begin{equation}
    \mathbf{M} = \begin{bmatrix} \cos\varphi & -\sin\varphi & t_x \\ \sin\varphi & \cos\varphi & t_y \end{bmatrix},
\end{equation}
with translation offsets $t_x = \frac{W'}{2} - \frac{W}{2}$ and $t_y = \frac{H'}{2} - \frac{H}{2}$ mapping to the expanded canvas of size $W' \times H'$, is applied jointly to the image and all surviving word bounding boxes. Each refined bbox is first expressed as a four-corner polygon $C_i = (x_1, y_1,\; x_2, y_1,\; x_2, y_2,\; x_1, y_2)$, and the corners are transformed under $\mathbf{M}$ to get the final 8-point polygon annotation format used for oriented text detection evaluation:
\begin{equation}
\hat{C}_i =
\left(
\begin{array}{c}
\mathbf{M}[\mathbf{p}_1^\top;1]^\top,\;
\mathbf{M}[\mathbf{p}_2^\top;1]^\top,\\
\mathbf{M}[\mathbf{p}_3^\top;1]^\top,\;
\mathbf{M}[\mathbf{p}_4^\top;1]^\top
\end{array}
\right)
\end{equation}
All annotations are stored in the format \texttt{word\ $x_0$\,$y_0$\ $x_1$\,$y_1$\ $x_2$\,$y_2$\ $x_3$\,$y_3$\ font\ class}, compatible with standard oriented document-text detection/recognition, layout and other benchmarks.

\section{Method Details}
\label{supp:method-details}

\subsection{Occlusion Detection and Blank Region Extraction}

Occlusion patches are first localized using a fine-tuned YOLOv9c detector \cite{wang2024yolov9} trained on the OPRB dataset. The benchmark contains six degradation classes, \emph{Black Ink}, \emph{Burnt}, \emph{Whitener}, \emph{Dust}, \emph{Scribble}, and \emph{Stamp}. Opaque classes (\emph{Black Ink}, \emph{Burnt}, \emph{Whitener}) fully obscure the underlying text, transparent classes (\emph{Dust}, \emph{Stamp}) allow partial visibility and no bounding boxes are omitted or trimmed, and the \emph{Scribble} class omits out individual words.

Once the YOLO detected patches $P = \{p_1, p_2, \ldots, p_k\}$ are detected (each $p_j$ given by absolute pixel coordinates $(x_1^j, y_1^j, x_2^j, y_2^j)$), blank regions are identified by a geometry-driven intersection method. For non-scribble patches, blanks are formed when a patch overlaps an inter-word gap, or when it forms a valid start-of-line or end-of-line blank candidate within the document text margins. Start/end blanks are accepted only after a three-sided enclosure check against neighbouring lines, duplicate emissions are suppressed, and every blank box is clamped to remain strictly within the detected occlusion patch. For scribble patches, words whose horizontal overlap with the patch exceeds 30\% of the word width are grouped into a contiguous run and converted into a single anchor-bounded blank.

We tokenize a recognized text line into three parts if and only if a blank is detected: \((i)\) the \emph{pre-text}, \((ii)\) the \emph{blank}, and \((iii)\) the \emph{post-text}. Let $G = \{w_0, w_1, \dots, w_n\}$ be a sorted group of word tokens for each line, where each token $w_j$ has an associated text field $\mathrm{text}(w_j)$. For a mid-line gap index $i$, we define
\[
\mathrm{PreText}_i \;=\; \mathrm{text}(w_0) \oplus \mathrm{text}(w_1) \oplus \cdots \oplus \mathrm{text}(w_{i}),
\]
\[
\mathrm{PostText}_i \;=\; \mathrm{text}(w_{i+1}) \oplus \mathrm{text}(w_{i+2}) \oplus \cdots \oplus \mathrm{text}(w_n),
\]
where $\oplus$ denotes string concatenation with spaces. For start-of-line blanks, $\mathrm{PreText}$ is empty and $\mathrm{PostText}$ is the full line and for end-of-line blanks, $\mathrm{PreText}$ is the full line and $\mathrm{PostText}$ is empty. Algorithm~\ref{alg:token_gen} constructs the formatted blank token written for each emitted gap. In RoBERTa mode, the same gap metadata are converted into per-blank masked queries for the language model.

\begin{algorithm}[t]
\caption{Prompt Token Generation for Emitted Blanks}
\label{alg:token_gen}
\begin{algorithmic}[1]
\Require A sorted line $G = \{w_0, w_1, \dots, w_n\}$, blank type $t \in \{\mathrm{start}, \mathrm{mid}, \mathrm{end}\}$, gap index $i$, blank width $b_w$, chars-per-pixel ratio $\rho$, blank id $\alpha$
\Ensure One formatted blank token and its metadata, or skip
\State $M \gets \max(1, \lfloor 1.2 \rho b_w \rfloor)$
\If{$M < 2$}
    \State \Return skip
\EndIf
\If{$t = \mathrm{start}$}
    \State $\mathrm{PreText}_\alpha \gets \texttt{""}$
    \State $\mathrm{PostText}_\alpha \gets \mathrm{text}(w_0) \oplus \cdots \oplus \mathrm{text}(w_n)$
\ElsIf{$t = \mathrm{mid}$}
    \State $\mathrm{PreText}_\alpha \gets \mathrm{text}(w_0) \oplus \cdots \oplus \mathrm{text}(w_i)$
    \State $\mathrm{PostText}_\alpha \gets \mathrm{text}(w_{i+1}) \oplus \cdots \oplus \mathrm{text}(w_n)$
\Else
    \State $\mathrm{PreText}_\alpha \gets \mathrm{text}(w_0) \oplus \cdots \oplus \mathrm{text}(w_n)$
    \State $\mathrm{PostText}_\alpha \gets \texttt{""}$
\EndIf
\State Construct $\mathrm{Token}_\alpha$ as
\[
\begin{aligned}
\mathrm{Token}_\alpha =\;&
\texttt{[K=}M_\alpha\texttt{] } \mathrm{PreText}_\alpha \;\texttt{<mask>}\; \mathrm{PostText}_\alpha
\end{aligned}
\]
\State \Return $\mathrm{Token}_\alpha$ and the corresponding blank metadata
\end{algorithmic}
\end{algorithm}


\subsection{Unified Restoration Details}

Our unified document restoration framework reconstructs degraded or occluded document images by first extracting occluded regions and then restoring them using a diffusion-based text editing model. In our implementation, the overall process consists of several stages: \((i)\) OCR and line grouping, \((ii)\) YOLO-based occlusion detection, \((iii)\) geometry-driven blank extraction when reconstruction is required, \((iv)\) missing-text prediction, \((v)\) adjacent-word inpainting and diffusion-based text editing, and \((vi)\) patch compositing followed by final occlusion-region cleanup.

First, our pipeline performs OCR on the input document image $I$ to detect text regions and generate a set of bounding boxes $Z_{bbox}$. The recognized text strings $S$ are then used to group the detected boxes into lines. For each line, blank candidates are extracted by intersecting YOLO patches with OCR word gaps, including start-of-line and end-of-line cases when a three-sided enclosure test is satisfied.

After the gaps are detected, our method creates tokenized representations of each gap by concatenating the recognized words before and after the gap. For semi-transparent occlusions (\emph{Dust} or \emph{Stamp}), the pipeline first checks whether sufficient OCR text remains visible inside the detected patches. If the text is sufficiently visible, tokenization and missing-text prediction are skipped and the method follows an inpaint-only branch over the intersecting word boxes. Otherwise, the newly formed gaps are sent to the fine-tuned RoBERTa model, which predicts the missing strings from context. The predicted target text dictionary $D$ is subsequently used in the text editing module, where a style patch $I_s$ is extracted from the original document. The text editing model then generates an edited patch $I_e$ conditioned on $I_s$ and $D_t$.

To ensure consistency between the restored document and the original, a postprocessing step is performed after text editing. First, a text inpainting module \cite{zhu2024text} is applied on the adjacent word bounding boxes of the detected blank to de-occlude the visible text and cleanup minor blobs and remains of the occluding patches. The edited patches are then cleaned and integrated back into the document, after which a final occlusion-region cleaning step removes residual borders, whitespace, and inter-line artifacts. The final restored document $I_{\text{final}}$ is obtained by placing the edited patch $I_e$ over the original image $I$ while also cleaning the gaps between nearby lines for visually seamless compositing.

\begin{figure*}[htbp]
  \centering
  \includegraphics[width=\textwidth]{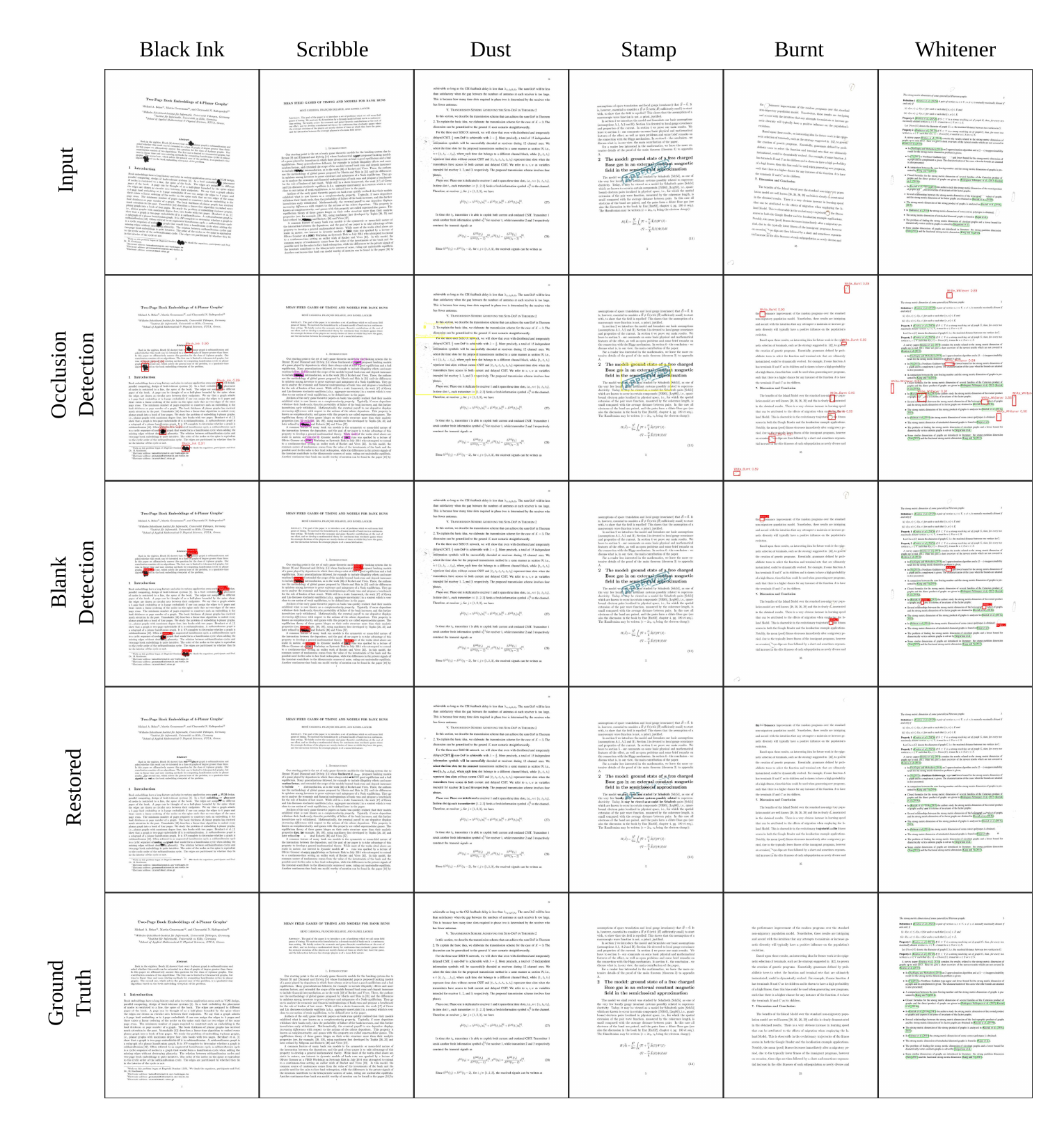}
  \caption{Visualization of the proposed framework on a few OPRB document images for different types of occlusions. [Zoom in for better visualization]}
  \label{fig:full-viz}
\end{figure*}

Figure~\ref{fig:full-viz} presents qualitative results of DocRevive across different occlusion types. These examples show that our method can accurately localize occluded regions, recover the missing text, and restore the document while preserving the original layout, spacing, typographic structure, and overall page appearance. Unlike generic image editing approaches, the restoration remains constrained to the damaged regions, which helps maintain both structural coherence and semantic integrity. These results suggest that our method is highly optimal and an effective solution for this challenging document restoration setting.

\begin{figure*}[htbp]
  \centering
  \includegraphics[width=\textwidth]{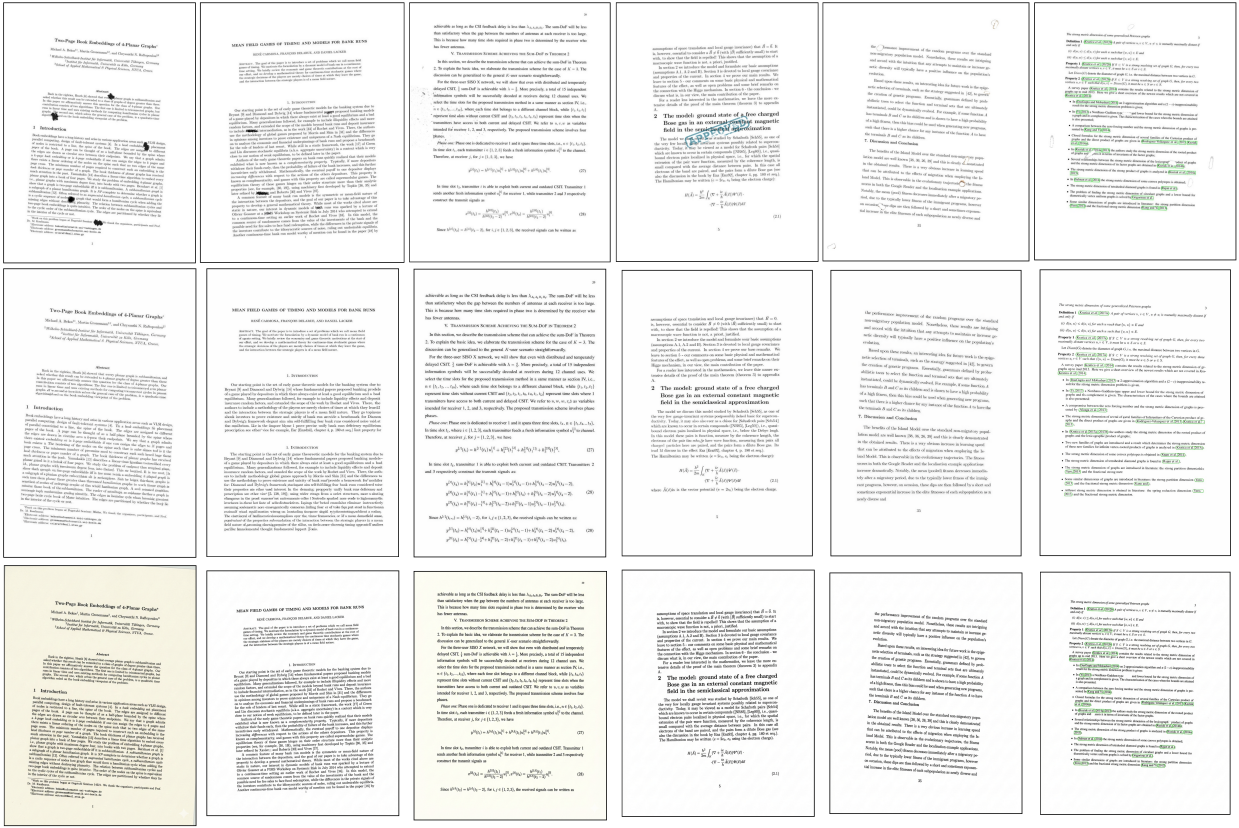}
  \caption{Visualization of the state of the art industry models on a few OPRB document images for different types of occlusions. [Zoom in for better visualization]}
  \label{fig:sota-viz}
\end{figure*}

Figure~\ref{fig:sota-viz} compares our task with state-of-the-art industry VLMs on the same occluded documents. ChatGPT shows substantial hallucinated content and often generates random text or repeated text with complete jargon within the document body, which severely harms document integrity. Nano Banana produces fewer hallucinations than ChatGPT, but it frequently alters the global document aesthetics, it often introduces a yellowish tint, edits irrelevant regions, and in several cases blurs the missing text or replaces it with a plain white blank. In contrast, our method preserves the original document style while restoring only the intended occluded content, thereby maintaining both structural consistency and semantic integrity. Since this is still a new problem setting, further research can continue improving restoration quality and robustness.

\subsection{Detailed UCSM Formulation}
\label{supp:ucsm-details}

We define UCSM as \emph{Unified Context Similarity Metric} for a predicted string $P$ and the corresponding ground-truth string $GT$. The formulation explicitly combines intrinsic similarity between $P$ and $GT$ with contextual difficulty derived from the surrounding text.

We denote the Levenshtein distance between $GT$ and $P$ by $d_{\text{lev}}(GT,P)$, and let $|GT|$ and $|P|$ denote the lengths (in characters) of the two strings. The normalized edit similarity is
\begin{equation}
S_{\text{edit}}(P,GT) = 1 - \frac{d_{\text{lev}}(GT,P)}{\max(|GT|,|P|)}.
\label{eq:edit_similarity}
\end{equation}
This score is $1$ when $P$ exactly matches $GT$, and it decreases as more edits are required to transform $P$ into $GT$.

To capture semantic similarity, we use cosine similarity over pretrained text embeddings. Let $\mathbf{e}(x)$ denote the embedding of a string $x$. Since cosine similarity lies in $[-1,1]$, we rescale it to $[0,1]$ as
\begin{equation}
S_{\text{sem}}(P,GT) = \frac{1}{2}\Bigl(\text{cosine}(\mathbf{e}(GT),\mathbf{e}(P)) + 1 \Bigr).
\label{eq:semantic_similarity}
\end{equation}

We also define a length-consistency term
\begin{equation}
S_{\text{len}}(P,GT) = \frac{\min(|GT|,|P|)}{\max(|GT|,|P|)}.
\label{eq:length_similarity}
\end{equation}
This term equals $1$ only when the two strings have identical length.

The three similarity terms are combined through a weighted geometric mean:
\begin{equation}
\begin{aligned}
S(P,GT) =
\Big[ &S_{\text{edit}}(P,GT)^\alpha \cdot
S_{\text{sem}}(P,GT)^\beta \cdot \\
& S_{\text{len}}(P,GT)^\gamma
\Big]^{\frac{1}{\alpha+\beta+\gamma}}
\end{aligned}
\label{eq:combined_similarity}
\end{equation}
where $\alpha$, $\beta$, and $\gamma$ are nonnegative weights assigned to edit, semantic, and length similarity respectively. In our experiments we set $\alpha = \beta = \gamma = 1$. The geometric mean is useful here because it suppresses the combined score whenever one of the components fails badly, which is desirable for restoration quality assessment.

To incorporate contextual difficulty, suppose the preceding text is denoted by $\text{pre}$ and the succeeding text by $\text{post}$. Let $\log P(\text{pre}+GT+\text{post})$ denote the log probability of the full text including $GT$, and let $\log P(\text{pre}+\text{post})$ denote the log probability of the context without $GT$. We approximate the contextual probability of the ground truth by $\log(P(GT \mid \text{pre}, \text{post}))$. After introducing calibration constants $m$ and $M$, the normalized context error is
\begin{equation}
E_{\text{context}} = \frac{-\log P(GT \mid \text{pre},\text{post}) - m}{M - m},
\label{eq:context_error}
\end{equation}
with $E_{\text{context}}$ clipped to the interval $[0,1]$. A lower value of $E_{\text{context}}$ indicates that the missing span is highly predictable from context, whereas a value near $1$ indicates low predictability.

Finally, UCSM combines intrinsic similarity and contextual difficulty as
\begin{equation}
UCSM = S(P,GT)^{(1 - E_{\text{context}})}.
\label{eq:UCSM-supp}
\end{equation}
If $P$ exactly equals $GT$, then $S(P,GT)=1$ and UCSM is $1$ regardless of context. For imperfect predictions, deviations are penalized more strongly when the context makes the correct answer highly predictable, which is precisely the behaviour desired for missing-text restoration.

When no embedding model is used, $S_{\text{sem}}$ defaults to $0.5$, when no LM log-probability function is used, $E_{\text{context}}$ defaults to $0.5$. When the LM term is enabled, the default calibration uses $m=-2$ and $M=10$.

\begin{figure*}[htbp]
  \centering
  \includegraphics[width=\textwidth]{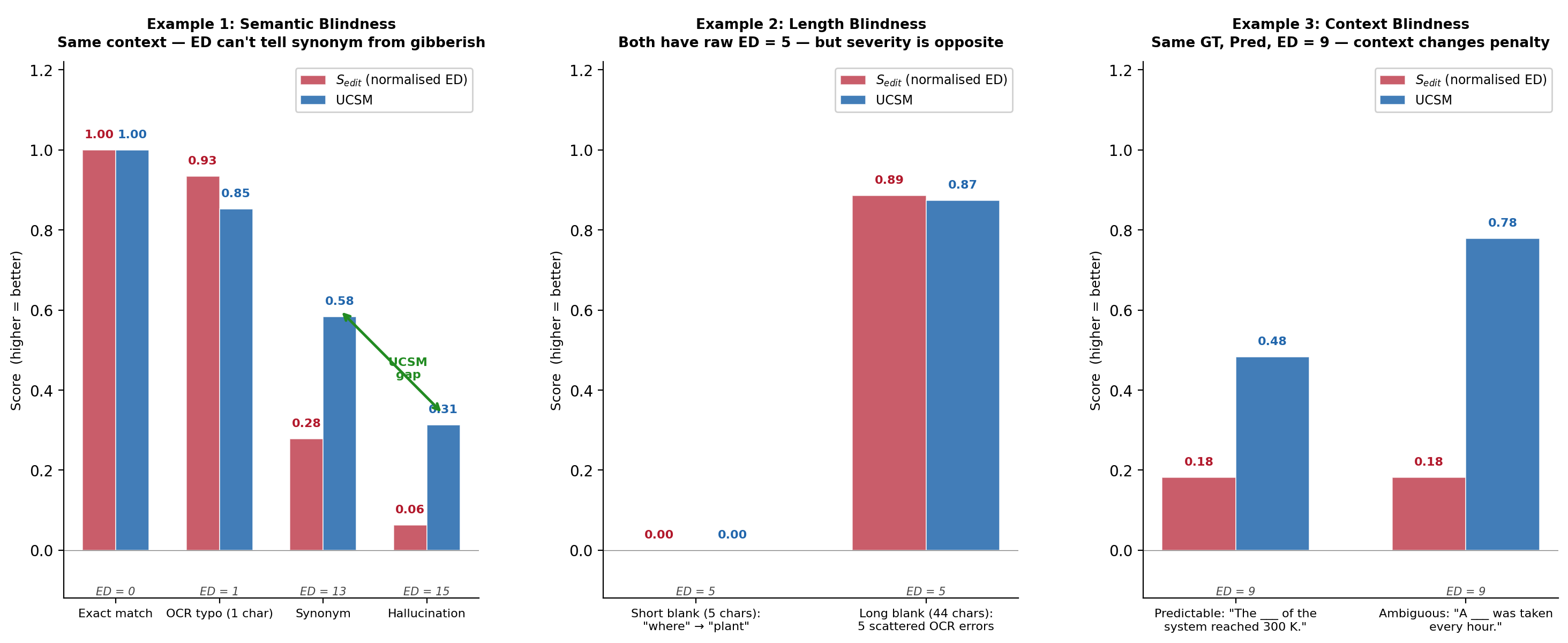}
  \caption{Graphical visualization of effectiveness of UCSM over Edit Distance [Zoom in for better visualization]}
  \label{fig:ucsmVed}
\end{figure*}

We perform a simple experiment to demonstrate how effective UCSM actually is. Table~\ref{tab:ucsm-examples} demonstrates three blind spots of Edit Distance (ED) that UCSM addresses through its multi-component design. In \textbf{Example~1}, we fix the context \emph{``We evaluate the \_\_\_ on three benchmark datasets.''} and ground truth \emph{``proposed method''}, then compare four predictions. ED ranks the valid synonym \emph{``suggested approach''} (ED\,=\,13) nearly as poorly as the hallucination \emph{``random variables''} (ED\,=\,15), whereas UCSM almost doubles the synonym's score (0.58 vs.\ 0.31) because $S_{\text{sem}}$ captures that the synonym preserves meaning. In \textbf{Example~2}, both predictions have exactly ED\,=\,5, yet one corrupts all 5 characters of a short word while the other introduces only 5 scattered OCR errors in a 44-character phrase. UCSM correctly assigns 0.00 to the total corruption and 0.87 to the minor error one, since $S_{\text{edit}}$ normalises by string length. In \textbf{Example~3}, the same wrong prediction \emph{``measurement''} for GT \emph{``temperature''} gives ED\,=\,9 in both cases, and $S_{\text{edit}}$, $S_{\text{sem}}$, and $S_{\text{len}}$ are identical. However, when the surrounding text is highly predictive (\emph{``\ldots reached 300\,K''}), $E_{\text{context}} \approx 0$ and UCSM penalises harshly (0.48), when the context is ambiguous (\emph{``A \_\_\_ was taken every hour''}), $E_{\text{context}} = 0.66$ and UCSM is more lenient (0.78). ED is entirely blind to this contextual distinction. The same is also presented in Figure~\ref{fig:ucsmVed}

\begin{table*}[t]
\centering
\small
\caption{Three blind spots of Edit Distance that UCSM addresses. $S_{\text{edit}}$, $S_{\text{sem}}$, and $S_{\text{len}}$ are defined in Eqs.~\eqref{eq:edit_similarity}--\eqref{eq:length_similarity}, $E_{\text{context}}$ in Eq.~\eqref{eq:context_error}, UCSM in Eq.~\eqref{eq:UCSM-supp}. Semantic embeddings are computed with all-MiniLM-L6-v2, context predictability with GPT-2.}
\label{tab:ucsm-examples}
\resizebox{\textwidth}{!}{%
\begin{tabular}{ll r cccc c}
\toprule
& \textbf{Prediction} & \textbf{ED} & $S_{\text{edit}}$ & $S_{\text{sem}}$ & $S_{\text{len}}$ & $E_{\text{ctx}}$ & \textbf{UCSM} \\
\midrule
\multicolumn{8}{l}{\textbf{Example 1: Semantic Blindness} \quad Context: \emph{``We evaluate the \underline{\hspace{1.2cm}} on three benchmark datasets.''} \quad GT: \emph{``proposed method''}} \\[3pt]
& \emph{proposed method} (exact match)       &  0 & 1.000 & 1.000 & 1.000 & 0.000 & \textbf{1.000} \\
& \emph{proposed methoc} (OCR typo)          &  1 & 0.933 & 0.665 & 1.000 & 0.000 & 0.853 \\
& \emph{suggested approach} (synonym)        & 13 & 0.278 & 0.859 & 0.833 & 0.000 & 0.584 \\
& \emph{random variables} (hallucination)    & 15 & 0.062 & 0.523 & 0.938 & 0.000 & 0.313 \\
\midrule
\multicolumn{8}{l}{\textbf{Example 2: Length Blindness} \quad Both predictions have ED\,=\,5, but severity is opposite} \\[3pt]
& \emph{plant} \scriptsize{(GT: ``where'', 5\,chars)}
                                              &  5 & 0.000 & 0.649 & 1.000 & 0.500 & \textbf{0.000} \\
& \scriptsize{5 OCR errors in 44-char phrase}
                                              &  5 & 0.886 & 0.503 & 1.000 & 0.500 & \textbf{0.874} \\
\midrule
\multicolumn{8}{l}{\textbf{Example 3: Context Blindness} \quad GT: \emph{``temperature''} $\to$ Pred: \emph{``measurement''} \quad (ED\,=\,9 in both cases)} \\[3pt]
& Predictable: \emph{``The \underline{\hspace{0.8cm}} of the system reached 300\,K.''}
                                              &  9 & 0.182 & 0.619 & 1.000 & 0.000 & \textbf{0.483} \\
& Ambiguous:\; \emph{``A \underline{\hspace{0.8cm}} was taken every hour.''}
                                              &  9 & 0.182 & 0.619 & 1.000 & 0.657 & \textbf{0.779} \\
\bottomrule
\end{tabular}
}
\end{table*}

\section{Misceleneous Experiments}
\subsection{Comparison with Prior Document Restoration Methods}
\label{supp:baseline-comparison}

We compare DocRevive against three prior methods on a subset of 498 images from OPRB (83 per occlusion type) DocDiff~\cite{yang2023docdiff}, GSDM (standalone), our pipeline's inpainting module run in isolation without any text prediction or editing and NAFNet~\cite{chen2022simple}, a strong general image restoration baseline adapted to document inputs. All methods receive the same degraded images. DocRevive however uses its full pipeline. Metrics are SSIM, PSNR, and FID against clean ground-truth pages. However, our pipeline cannot be evaluated against other document based or real based dataset as because no similar fully text occluded document dataset exists to the best of our knowledge. Therefore, only the result of different methods on our dataset is presented.

Table~\ref{tab:baseline-overall} shows overall performance. DocRevive achieves the highest SSIM and PSNR, and dramatically outperforms all baselines on FID (9.00 vs.\ 43.25 for the next-best method). The FID gap is particularly telling that image-level similarity metrics such as SSIM and PSNR measure pixel correspondence, but FID captures perceptual distribution quality, where DocRevive's text-aware restoration produces outputs that are statistically far closer to real clean documents than any purely visual method.

\begin{table}[h]
\centering
\small
\caption{Overall comparison on the OPRB subset. Best results in \textbf{bold}. Second best in \underline{underline}.}
\label{tab:baseline-overall}
\begin{tabular}{lccc}
\toprule
\textbf{Method} & \textbf{SSIM\,$\uparrow$} & \textbf{PSNR\,$\uparrow$} & \textbf{FID\,$\downarrow$} \\
\midrule
DocDiff         & 0.6620 & 16.46 & \underline{43.25} \\
GSDM  & 0.6634 & 16.58 & 53.78 \\
NAFNet            & \underline{0.6663} & \textbf{17.42} & 67.57 \\
\midrule
\textbf{DocRevive (Ours)}     & \textbf{0.6814} & \underline{17.23} & \textbf{9.00} \\
\bottomrule
\end{tabular}
\end{table}

\paragraph{Per-occlusion-type breakdown.}
Tables~\ref{tab:baseline-ssim-psnr} report per-type SSIM and PSNR. DocRevive consistently leads across all occlusion types in SSIM and FID. The GSDM standalone baseline performs comparably to DocRevive on opaque patch types (Black Ink SSIM 0.6814 vs.\ 0.6816), confirming that pure inpainting suffices for opaque occlusions; the gap widens substantially on transparent classes (Dust, Stamp) and visually complex classes (Burnt, Whitener), where text-aware reconstruction is necessary. On PSNR, NAFNet is competitive on Dust and Whitener due to its effective denoising capability, yet DocRevive's semantic reconstruction gives substantially better perceptual quality as reflected by FID.

\begin{table*}[t]
\centering
\small
\caption{Per-occlusion-type SSIM\,$\uparrow$ and PSNR\,$\uparrow$ on the OPRB. Best in \textbf{bold}. Second best in \underline{underline}}
\label{tab:baseline-ssim-psnr}
\setlength{\tabcolsep}{5pt}
\begin{tabular}{lcccccc@{\hskip 12pt}cccccc}
\toprule
& \multicolumn{6}{c}{\textbf{SSIM\,$\uparrow$}} & \multicolumn{6}{c}{\textbf{PSNR\,$\uparrow$}} \\
\cmidrule(lr){2-7}\cmidrule(lr){8-13}
\textbf{Method}
  & \textbf{Blk.\,Ink} & \textbf{Scrib.} & \textbf{Dust} & \textbf{Stamp} & \textbf{Burnt} & \textbf{Whit.}
  & \textbf{Blk.\,Ink} & \textbf{Scrib.} & \textbf{Dust} & \textbf{Stamp} & \textbf{Burnt} & \textbf{Whit.} \\
\midrule
DocDiff
  & 0.6792 & 0.6726 & 0.6416 & 0.6538 & 0.6748 & 0.6500
  & 15.90  & 16.90  & 16.45  & 16.12  & 16.91  & 16.51 \\
GSDM
  & \underline{0.6814} & 0.\textbf{6751} & 0.6458 & 0.6556 & 0.6742 & 0.6486
  & 16.93  & \underline{17.37}  & 15.61  & 15.75  & 17.11  & 16.72 \\
NAFNet
  & 0.6790 & 0.6699 & \underline{0.6490} & \underline{0.6609} & \underline{0.6805} & \underline{0.6587}
  & 16.73  & \textbf{17.77}  & \textbf{17.44}  & \textbf{17.10}  & \textbf{17.88}  & \textbf{17.59} \\
\midrule
\textbf{DocRevive}
  & \textbf{0.6816} & \underline{0.6745} & \textbf{0.6833} & \textbf{0.6768} & \textbf{0.6866} & \textbf{0.6855}
  & \textbf{17.24} & 17.09 & \underline{17.26} & \underline{16.83} & \underline{17.48} & \underline{17.50} \\
\bottomrule
\end{tabular}
\end{table*}

\begin{figure*}[t]
    \centering
    \setlength{\tabcolsep}{1.5pt}
    \renewcommand{\arraystretch}{1}
    \setlength{\fboxrule}{0.25pt}
    \setlength{\fboxsep}{0pt}

    \newlength{\imgw}
    \newlength{\imgh}
    \setlength{\imgw}{0.147\textwidth} 
    \setlength{\imgh}{0.200\textwidth}
    \newcommand{\rlabel}[1]{\rotatebox{90}{\scriptsize #1}}

    \newcommand{\imgcell}[1]{%
        \fbox{\includegraphics[width=\imgw, height=\imgh]{#1}}%
    }

    \begin{tabular}{
        @{}
        >{\centering\arraybackslash}m{1.6em}
        @{\hspace{2pt}}
        *{6}{>{\centering\arraybackslash}m{\imgw}}
        @{}
    }

        &
        \scriptsize Black Ink &
        \scriptsize Scribble &
        \scriptsize Dust &
        \scriptsize Stamp &
        \scriptsize Burnt &
        \scriptsize Whitener
        \\[3pt]

        \rlabel{Input} &
        \imgcell{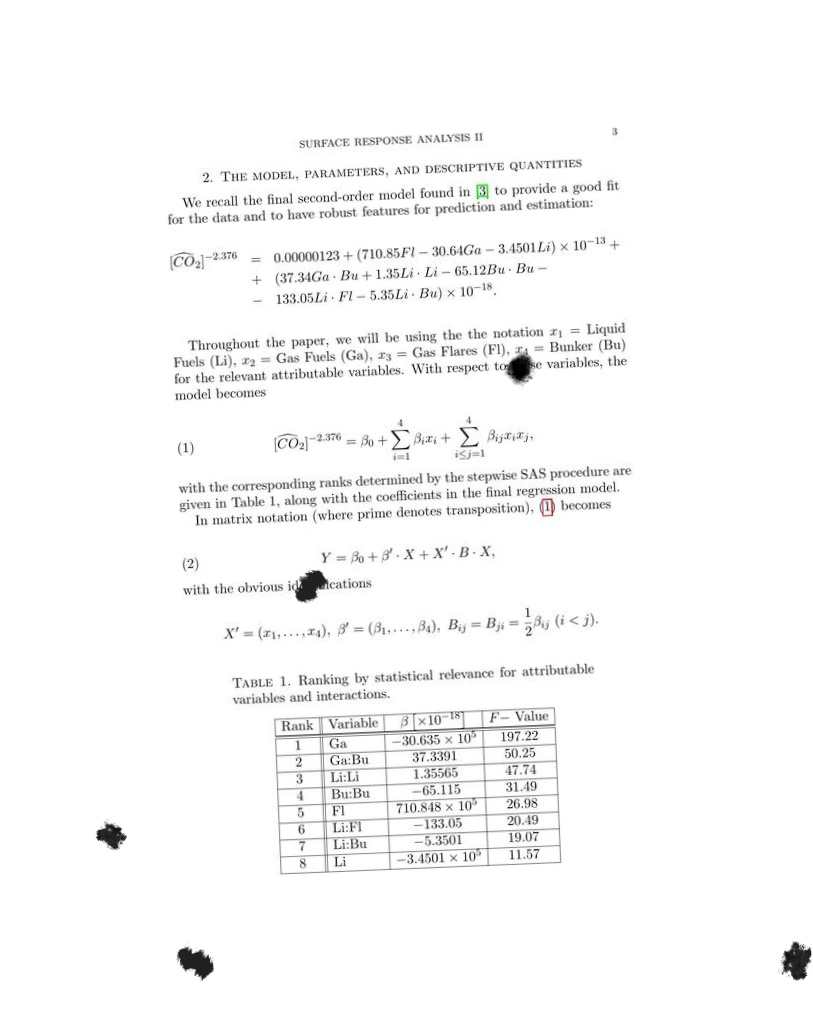} &
        \imgcell{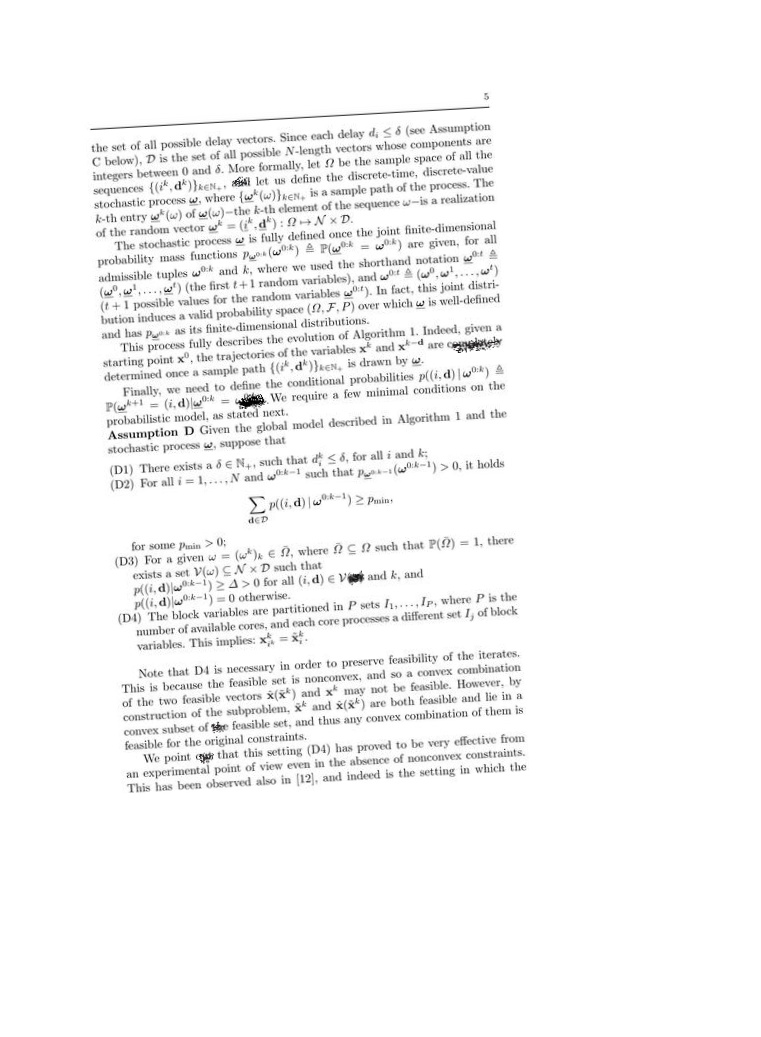} &
        \imgcell{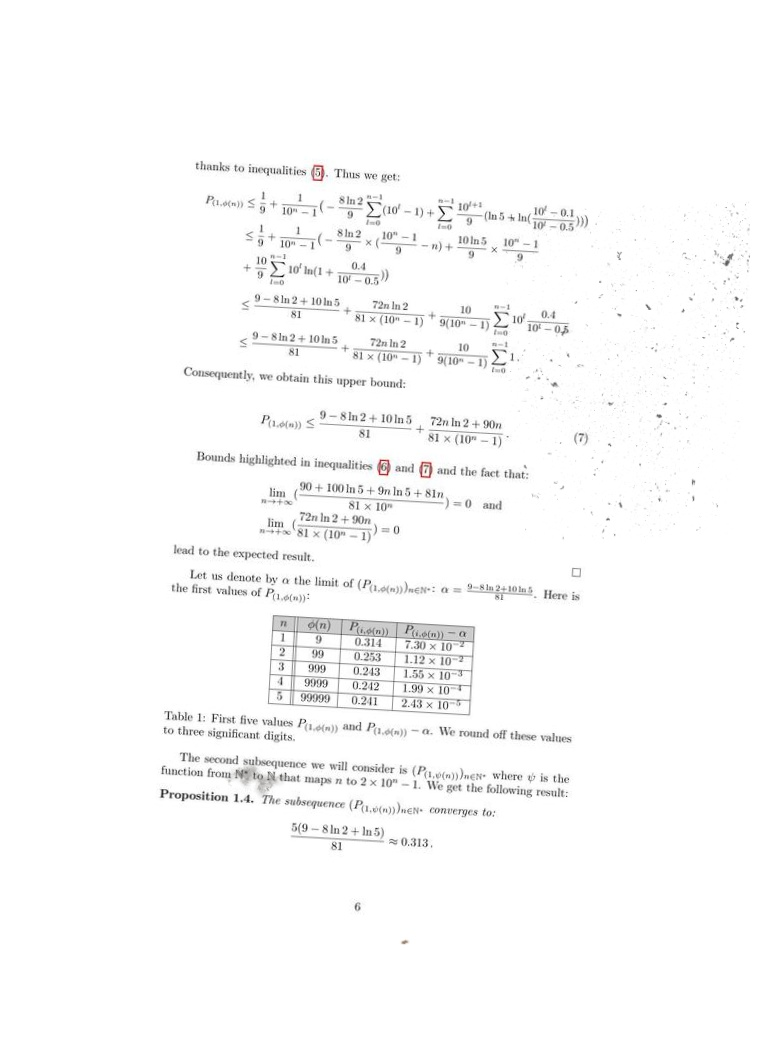} &
        \imgcell{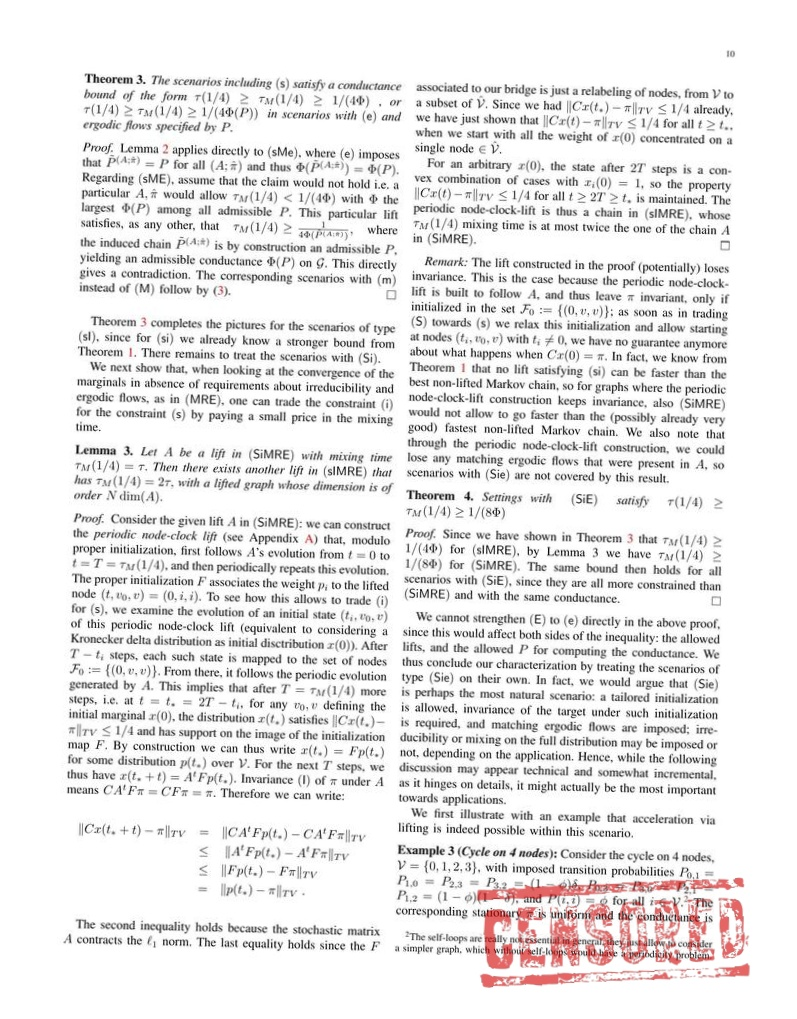} &
        \imgcell{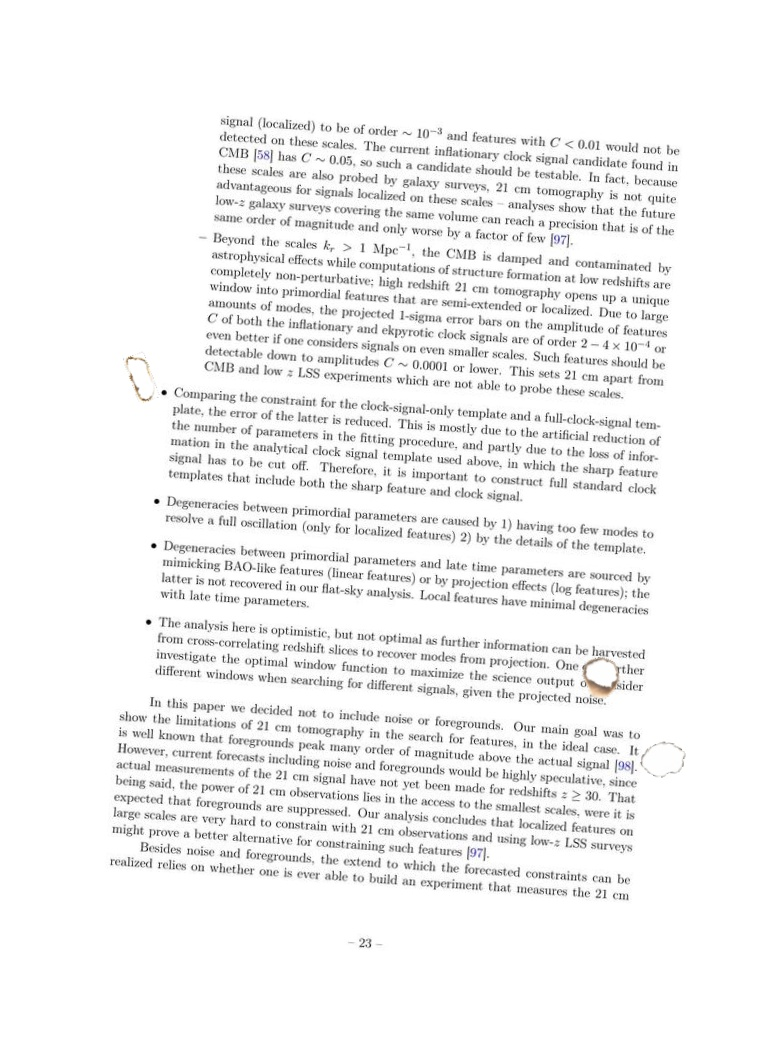} &
        \imgcell{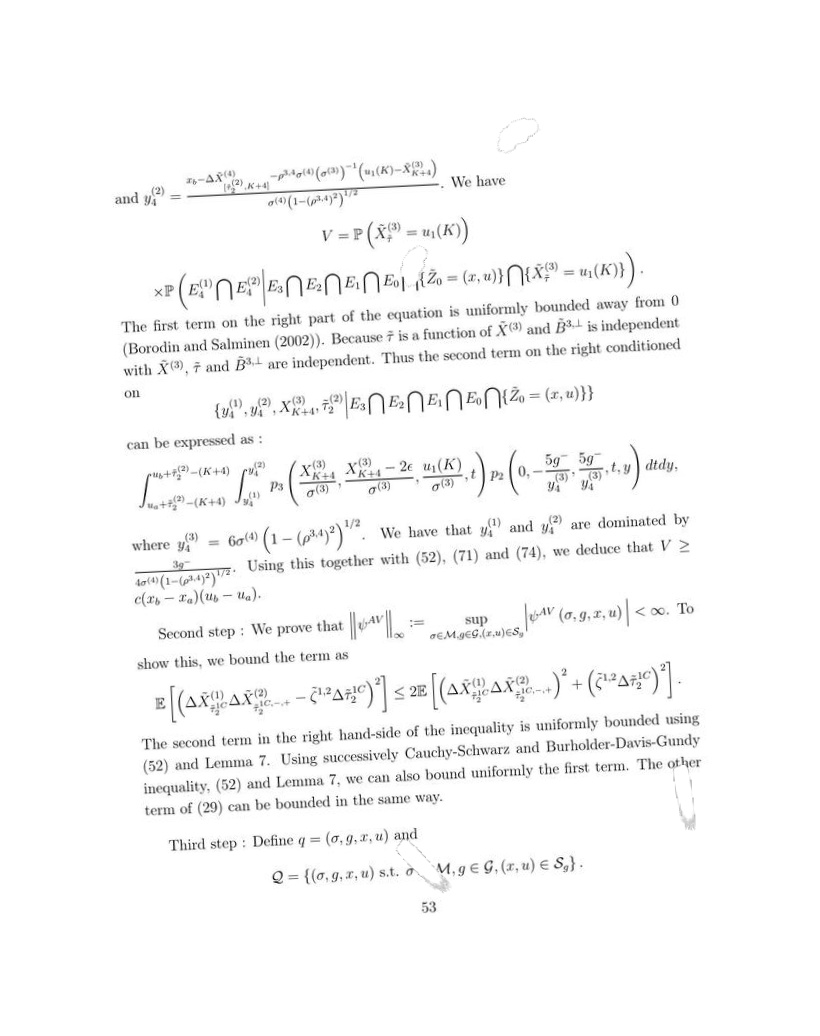} \\

        \rlabel{NAFNet} &
        \imgcell{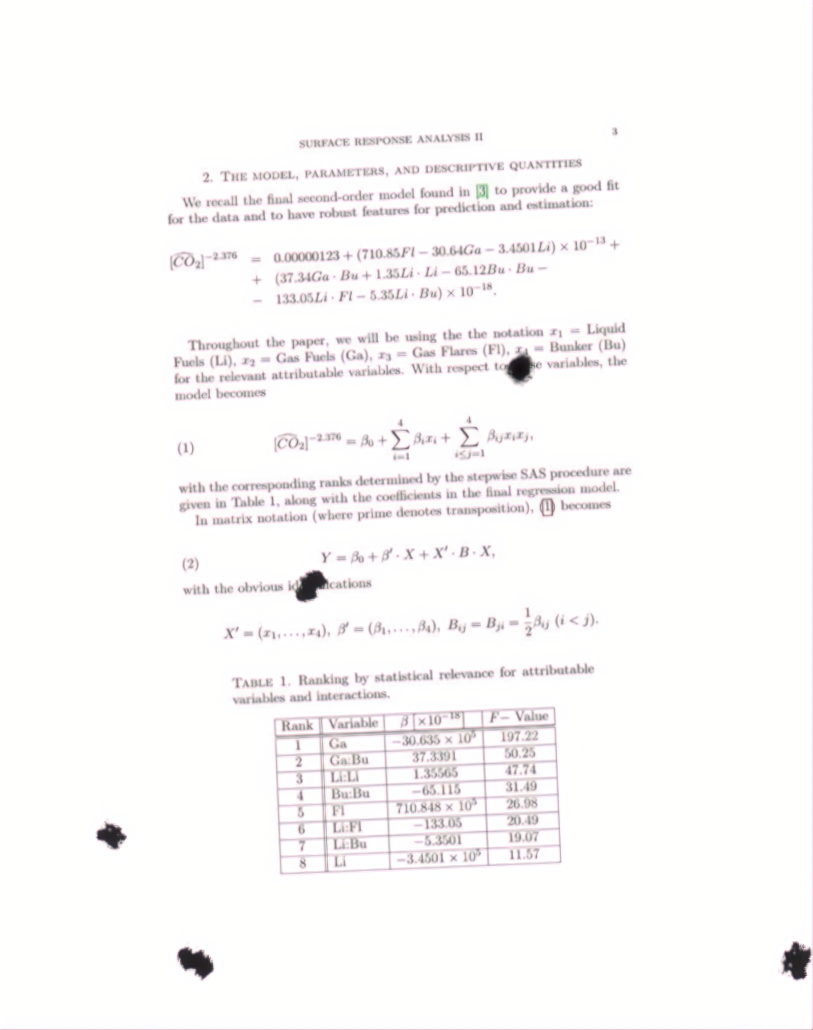} &
        \imgcell{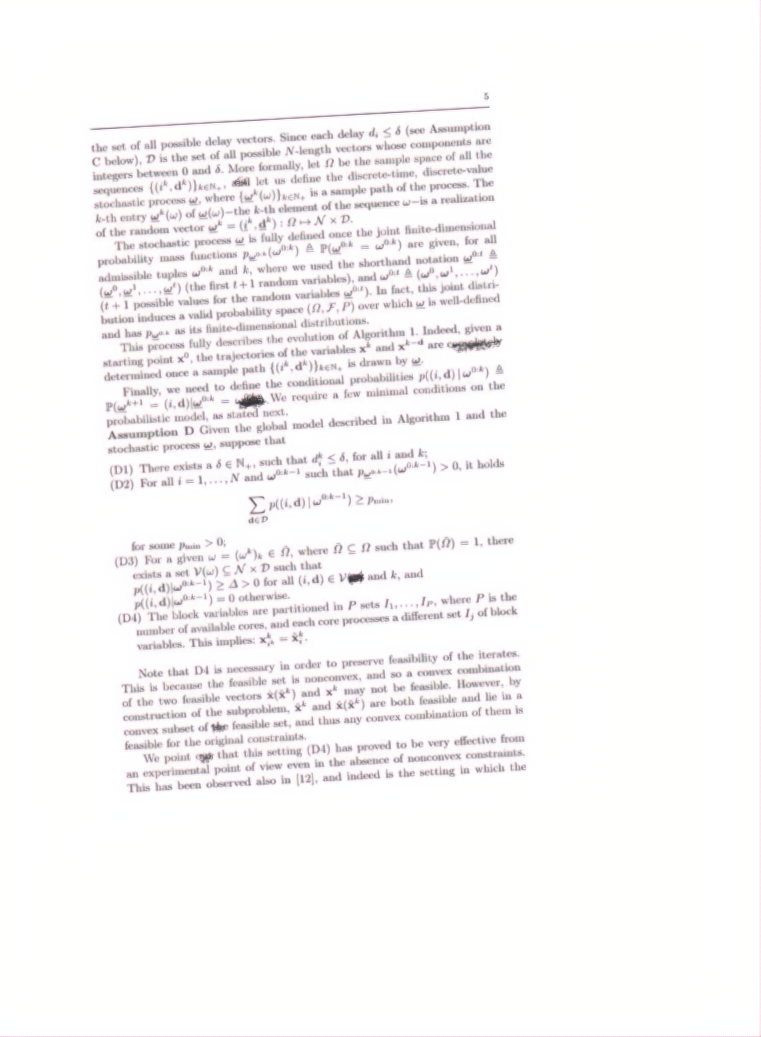} &
        \imgcell{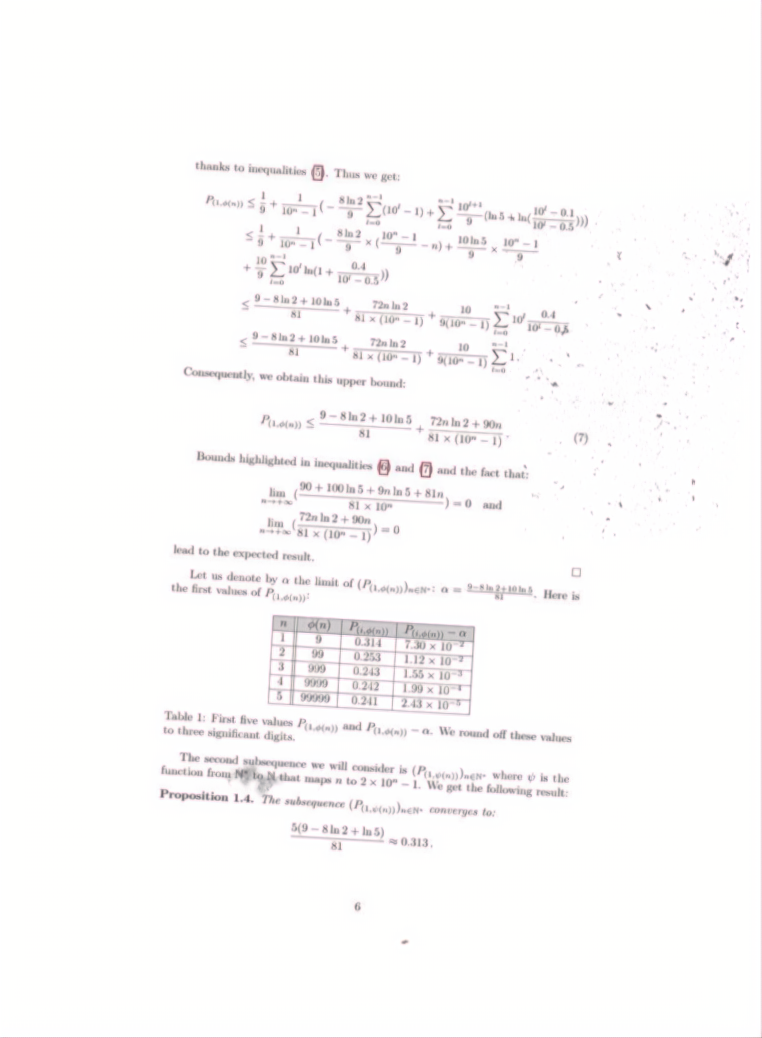} &
        \imgcell{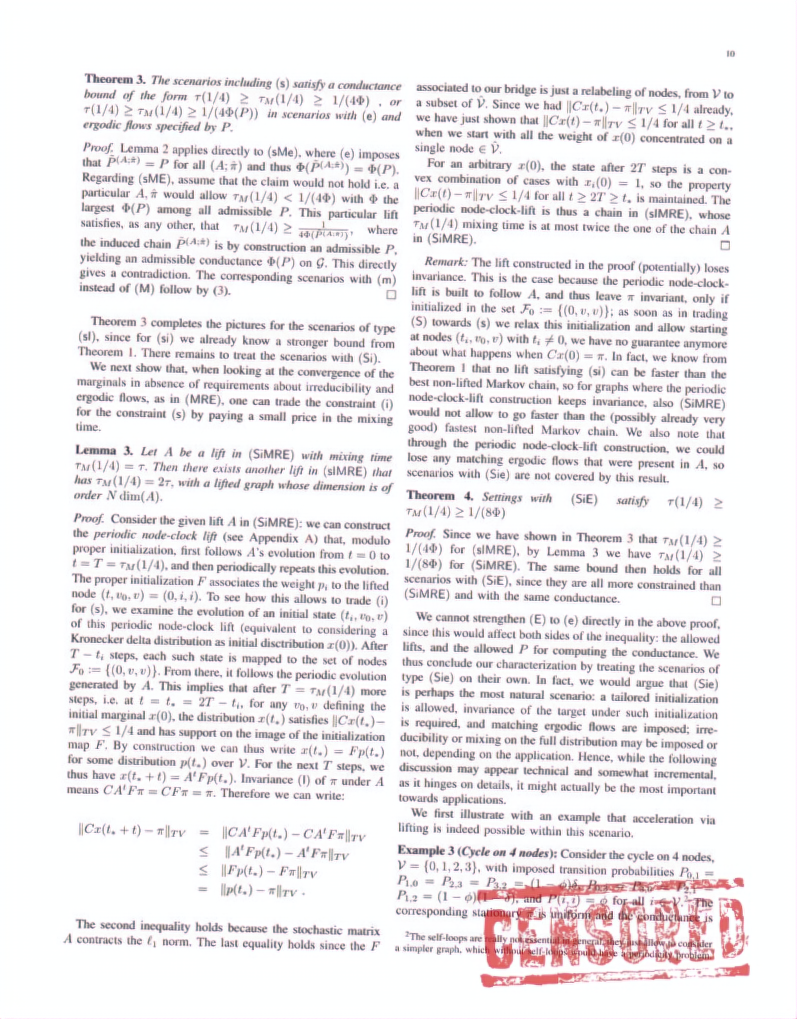} &
        \imgcell{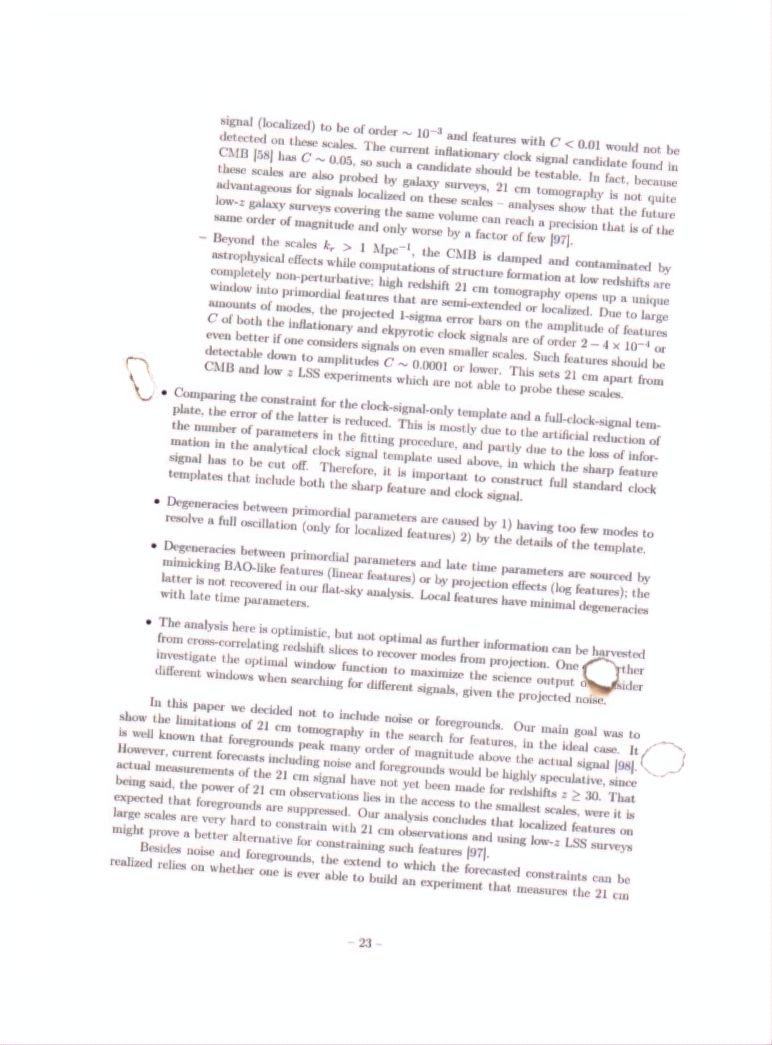} &
        \imgcell{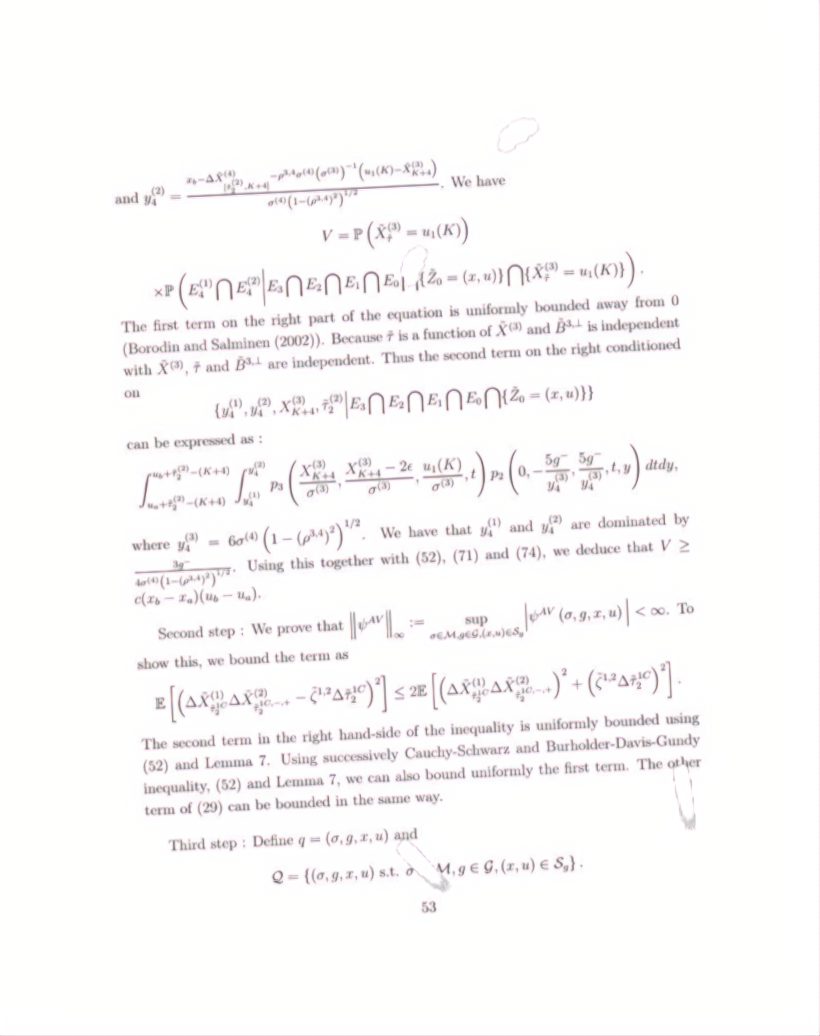} \\

        \rlabel{DocDiff} &
        \imgcell{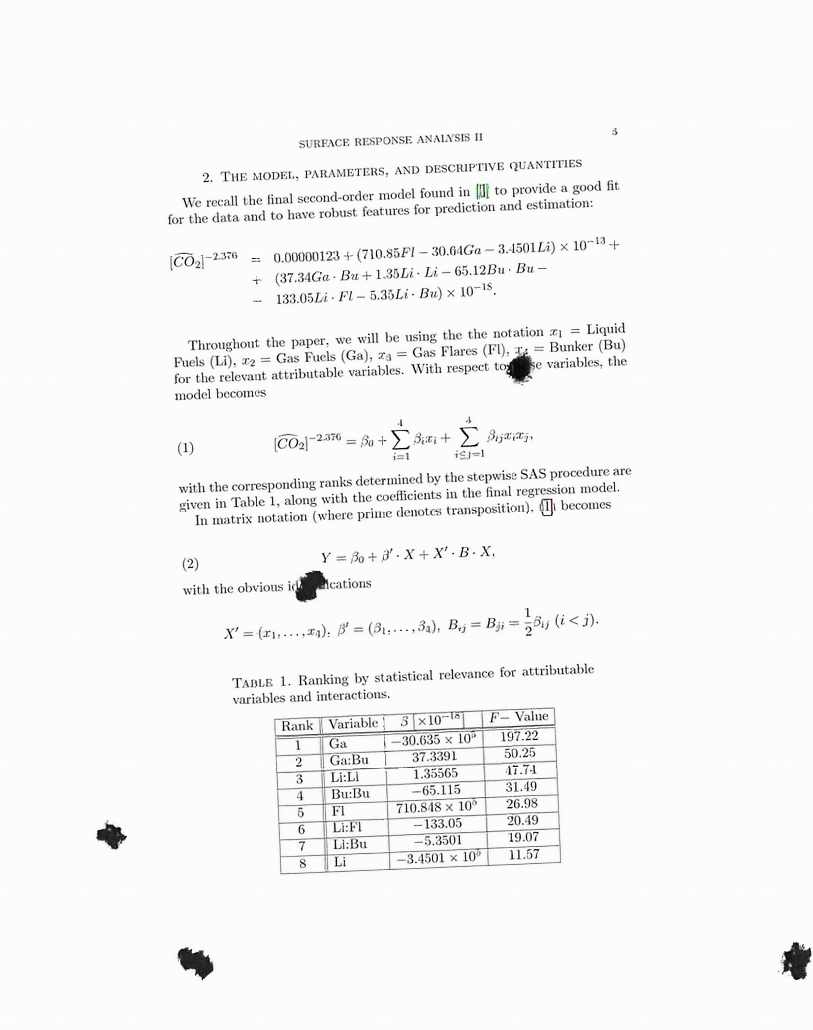} &
        \imgcell{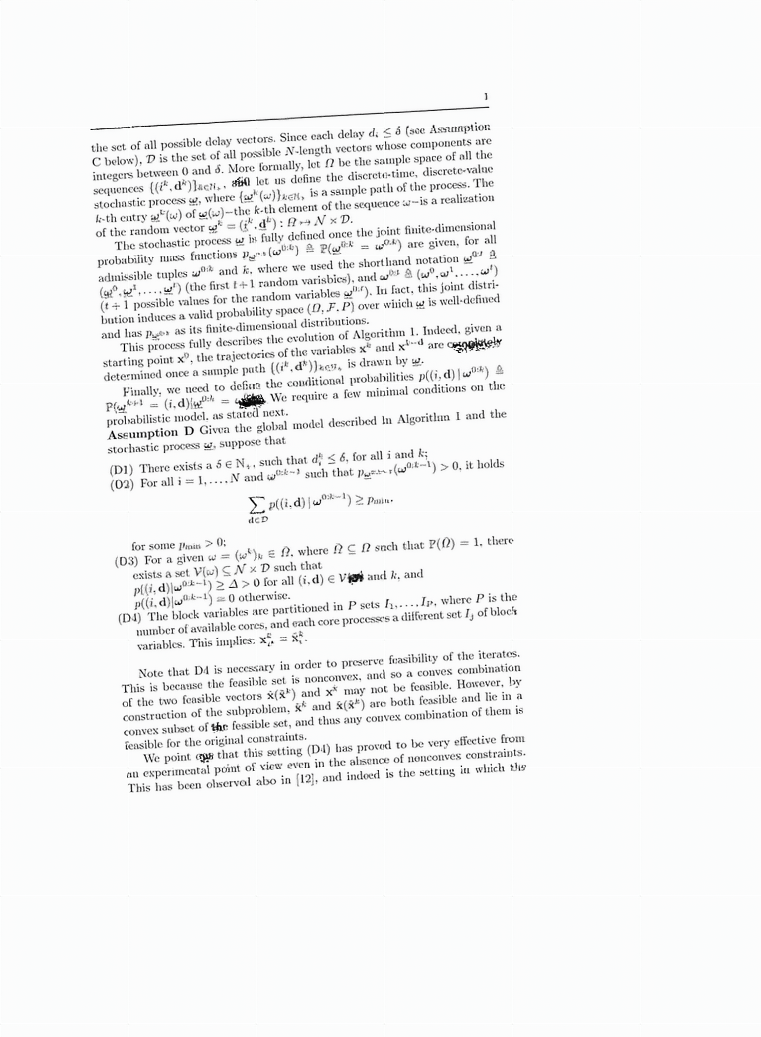} &
        \imgcell{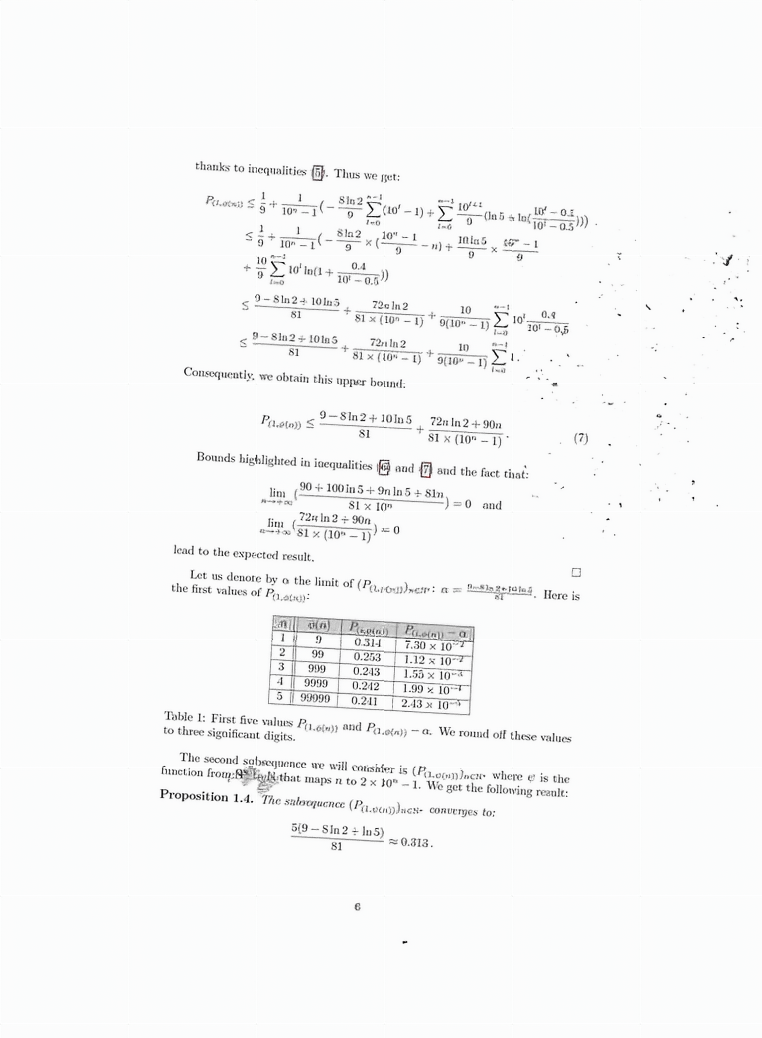} &
        \imgcell{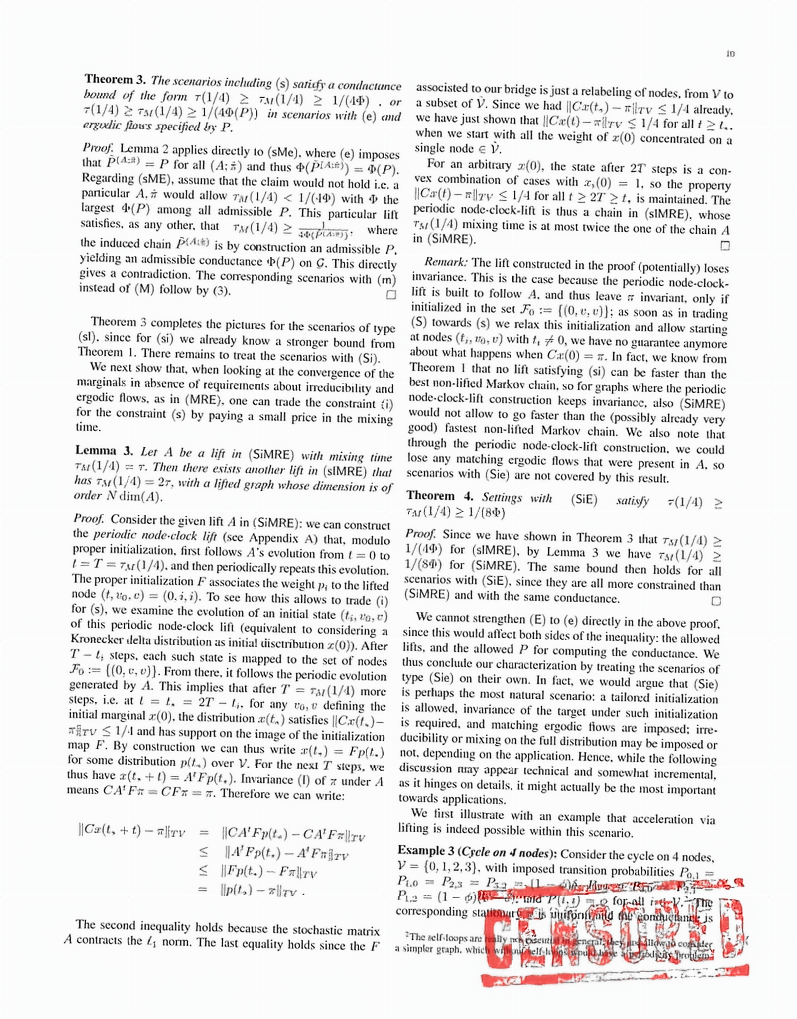} &
        \imgcell{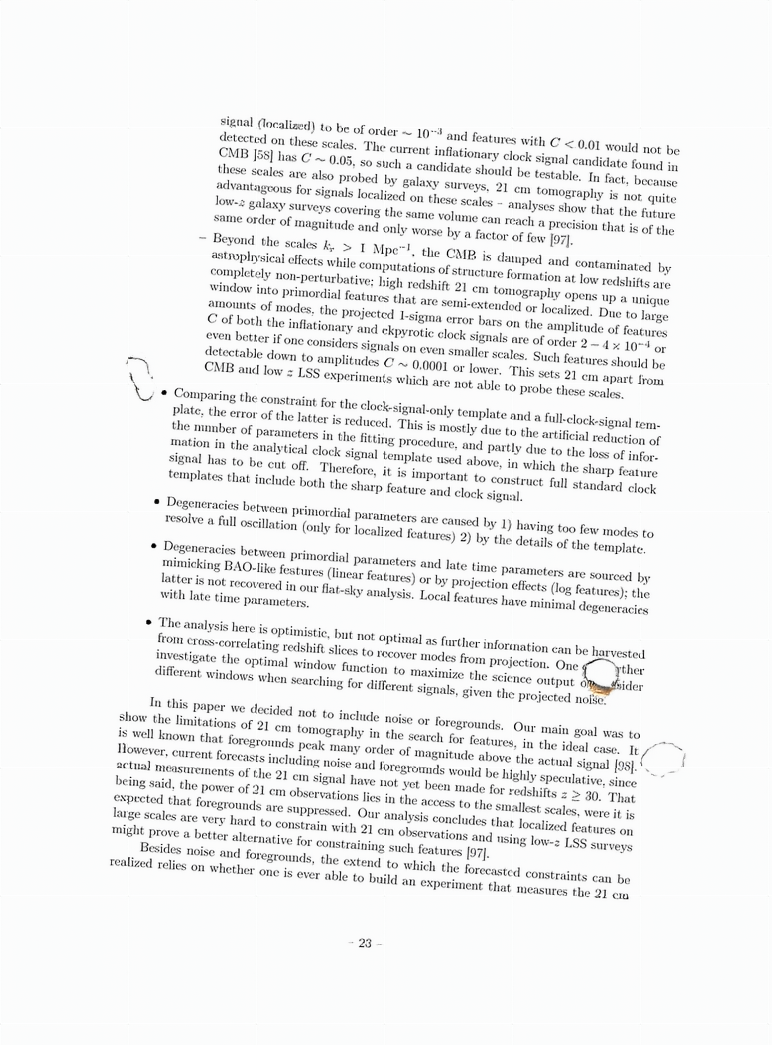} &
        \imgcell{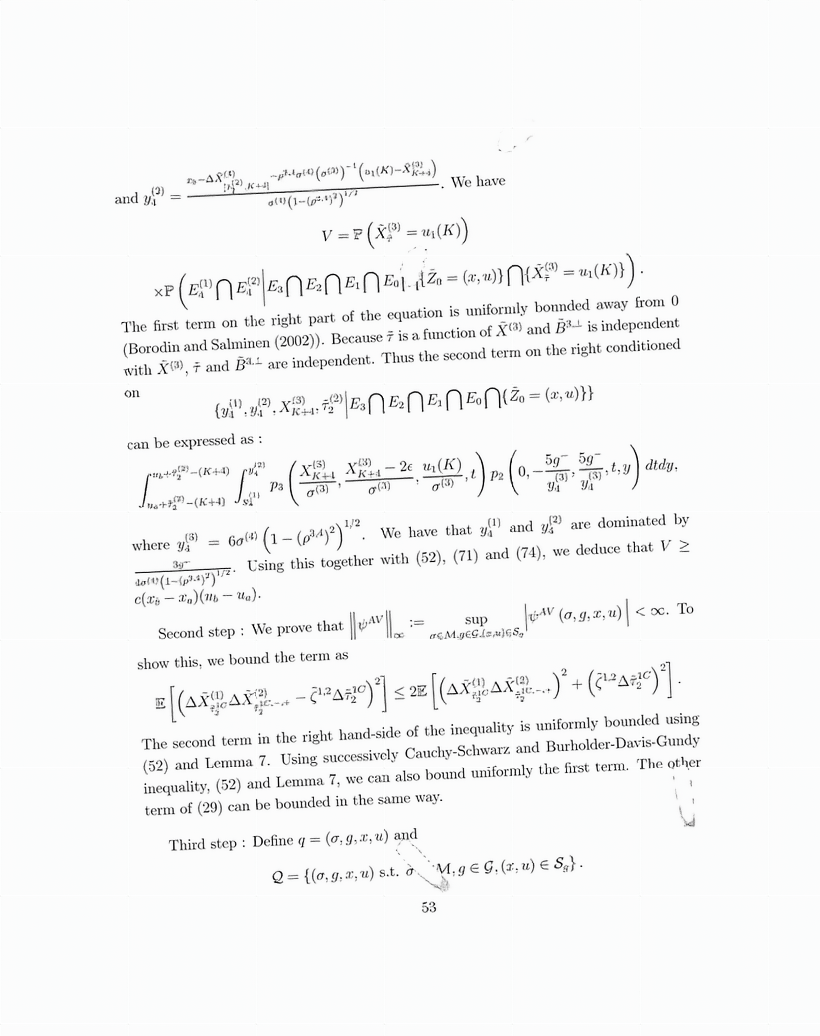} \\

        \rlabel{GSDM} &
        \imgcell{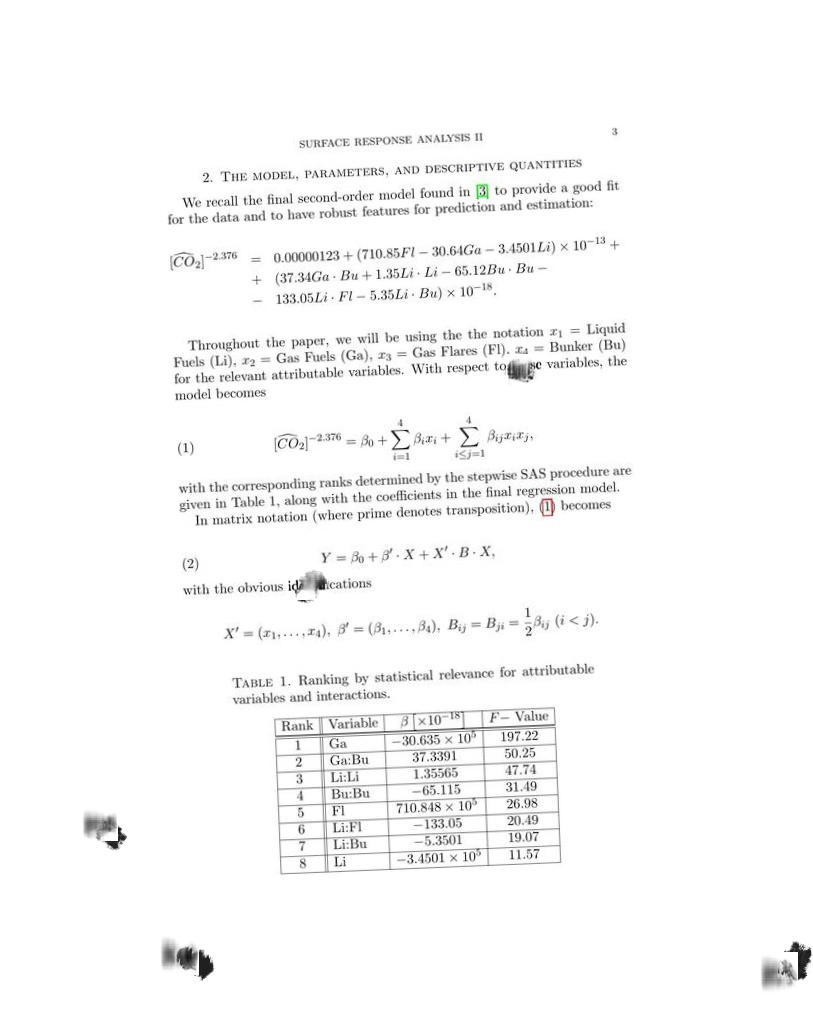} &
        \imgcell{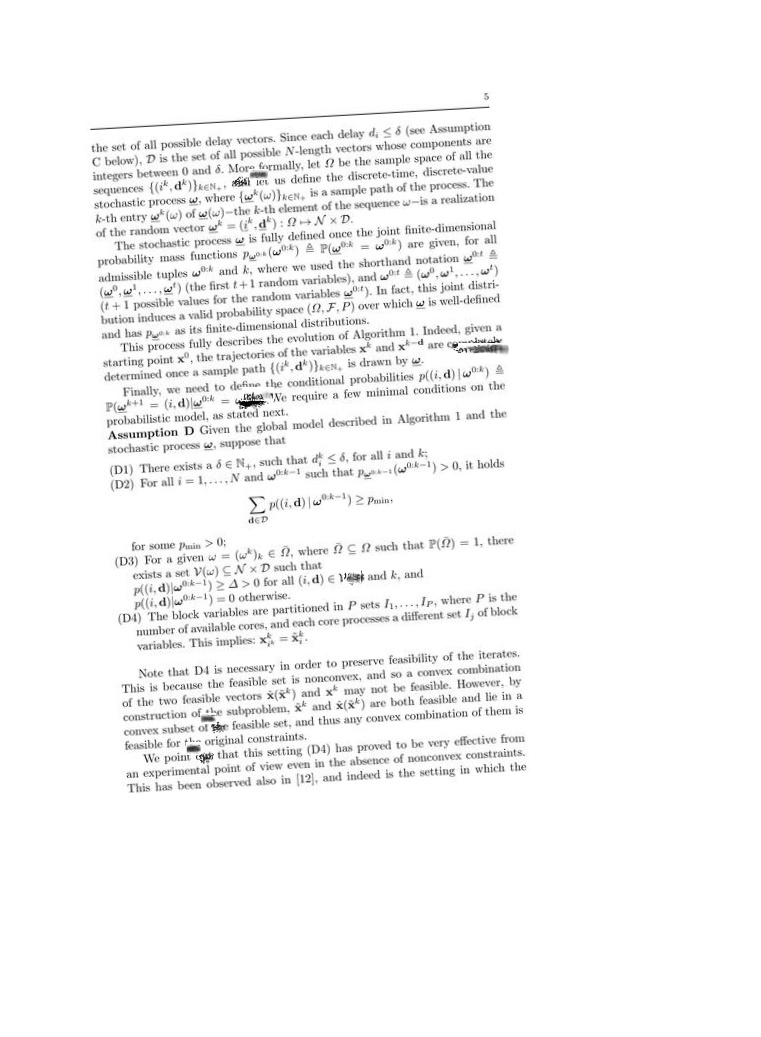} &
        \imgcell{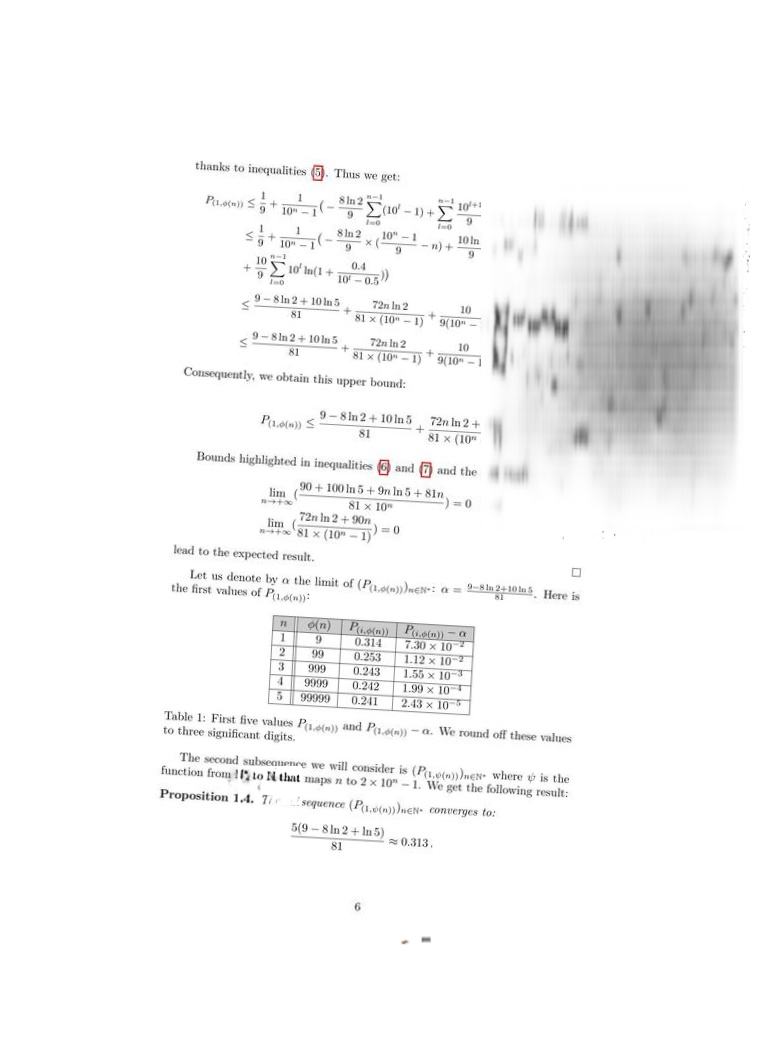} &
        \imgcell{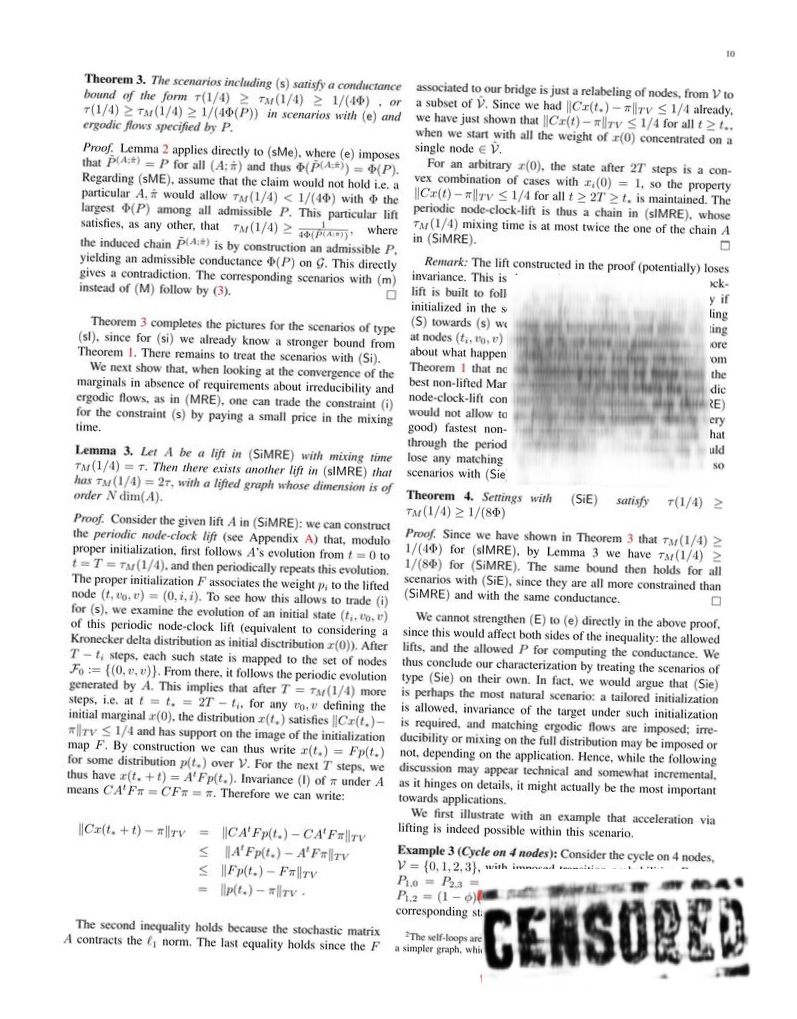} &
        \imgcell{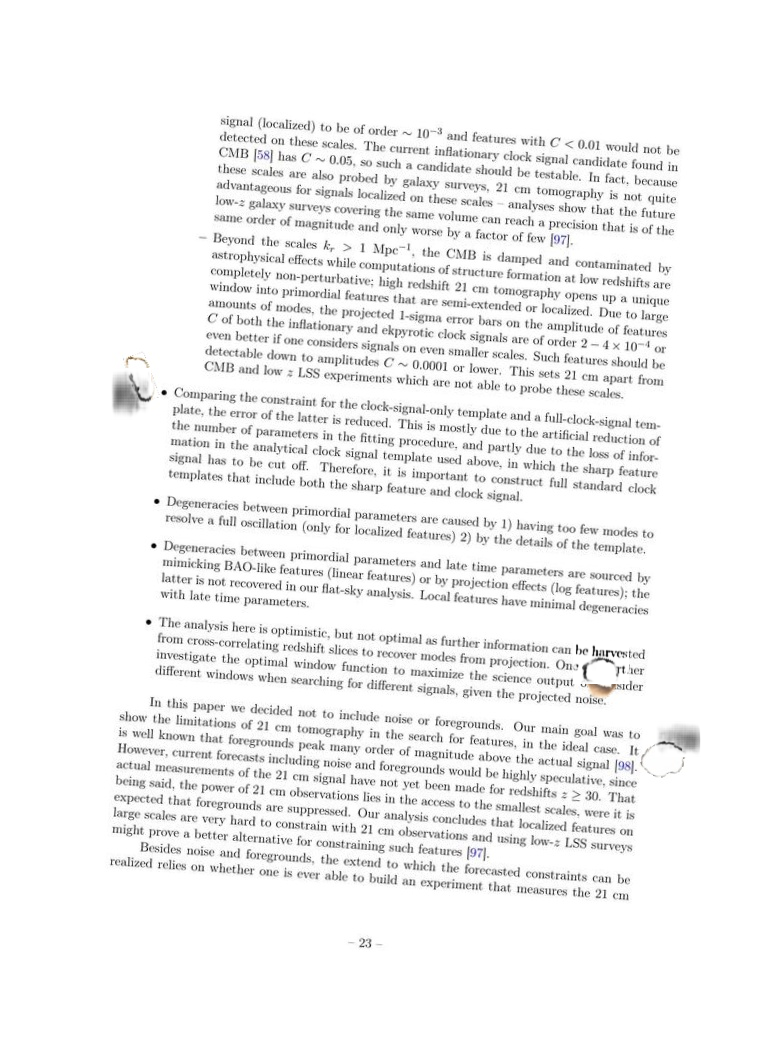} &
        \imgcell{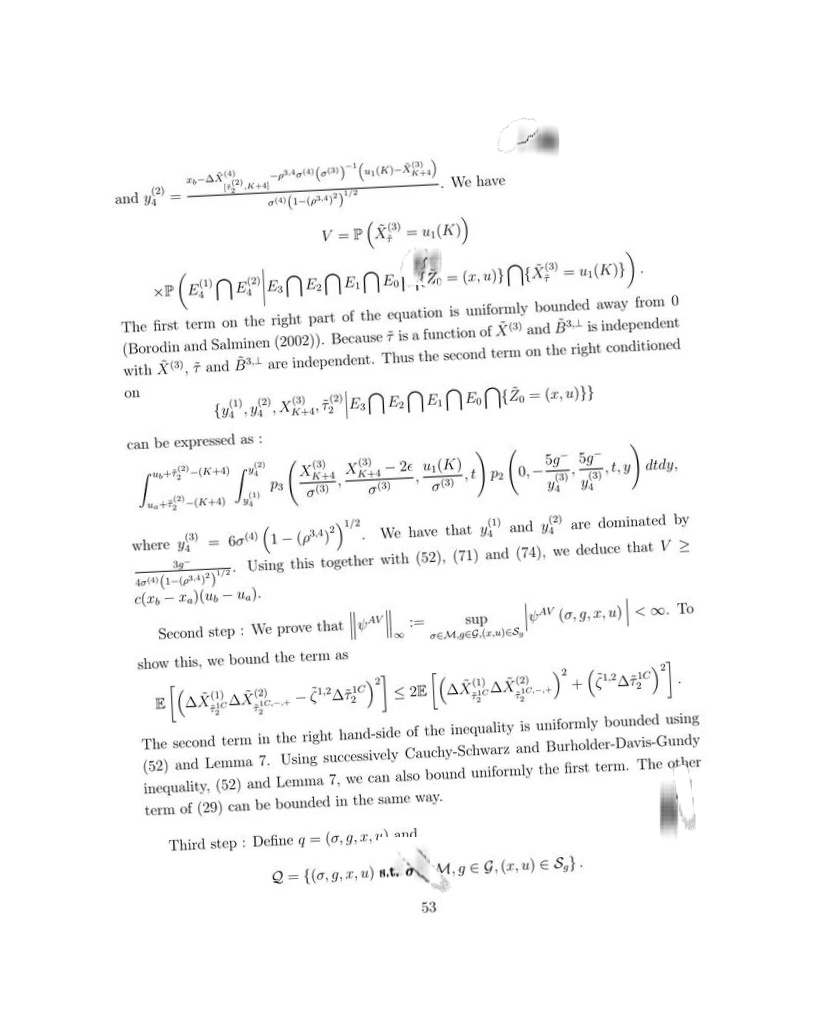} \\

        \rlabel{DocRevive} &
        \imgcell{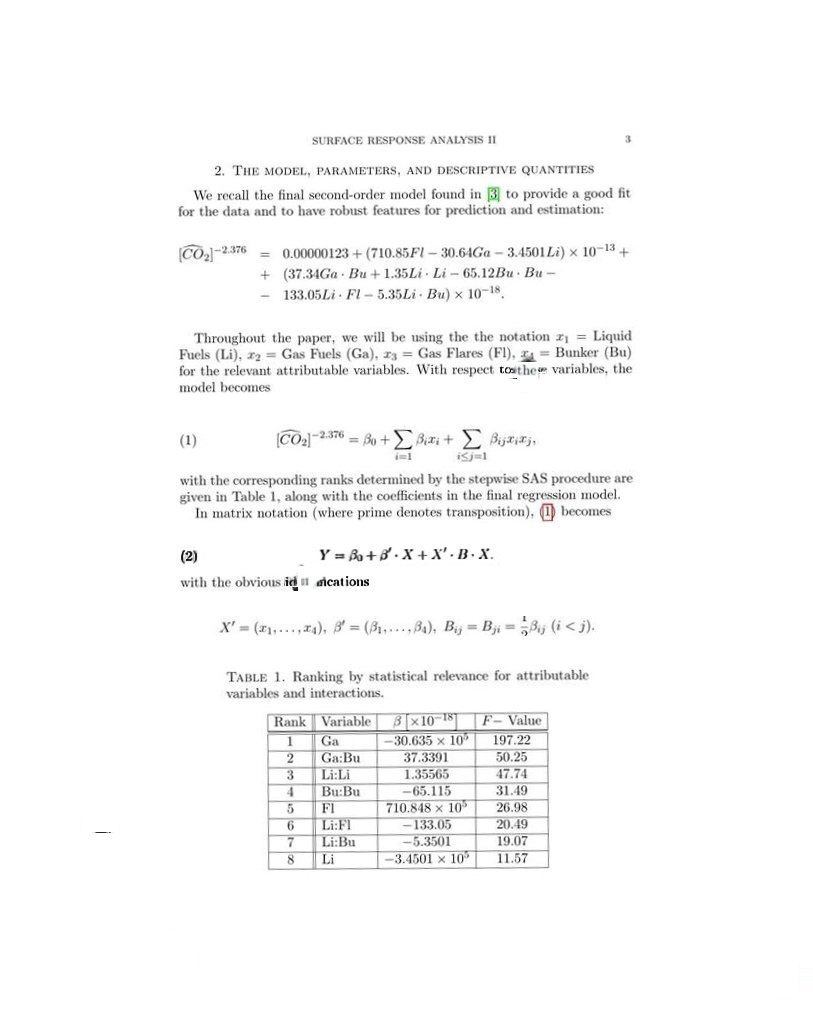} &
        \imgcell{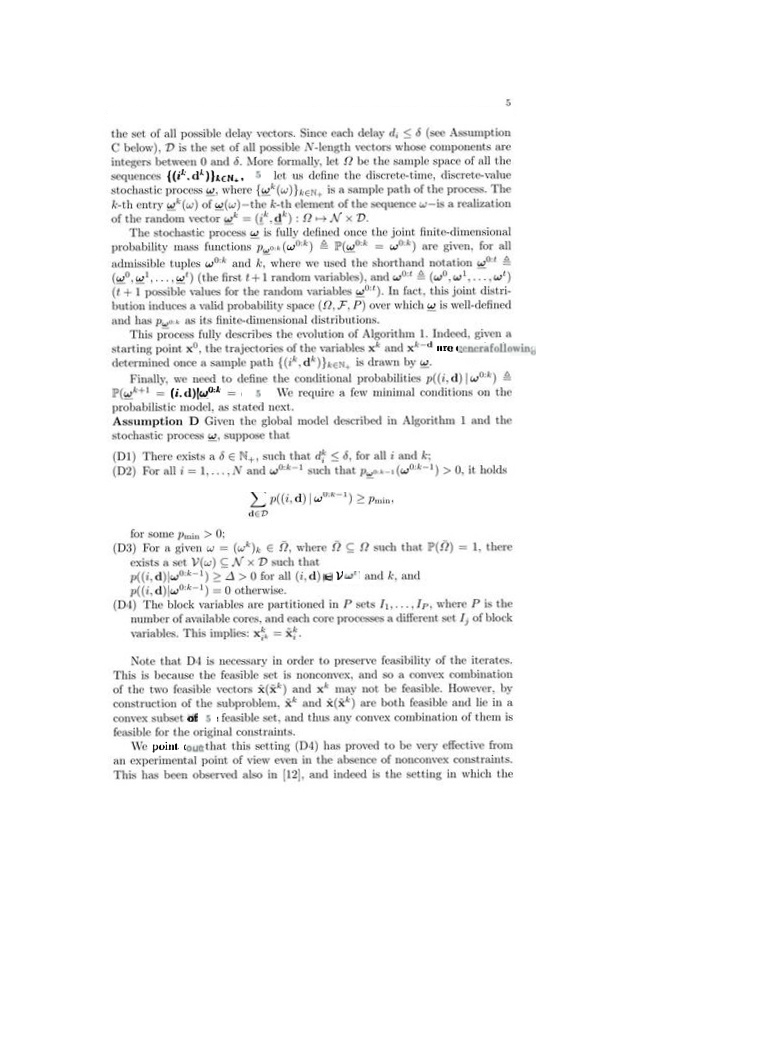} &
        \imgcell{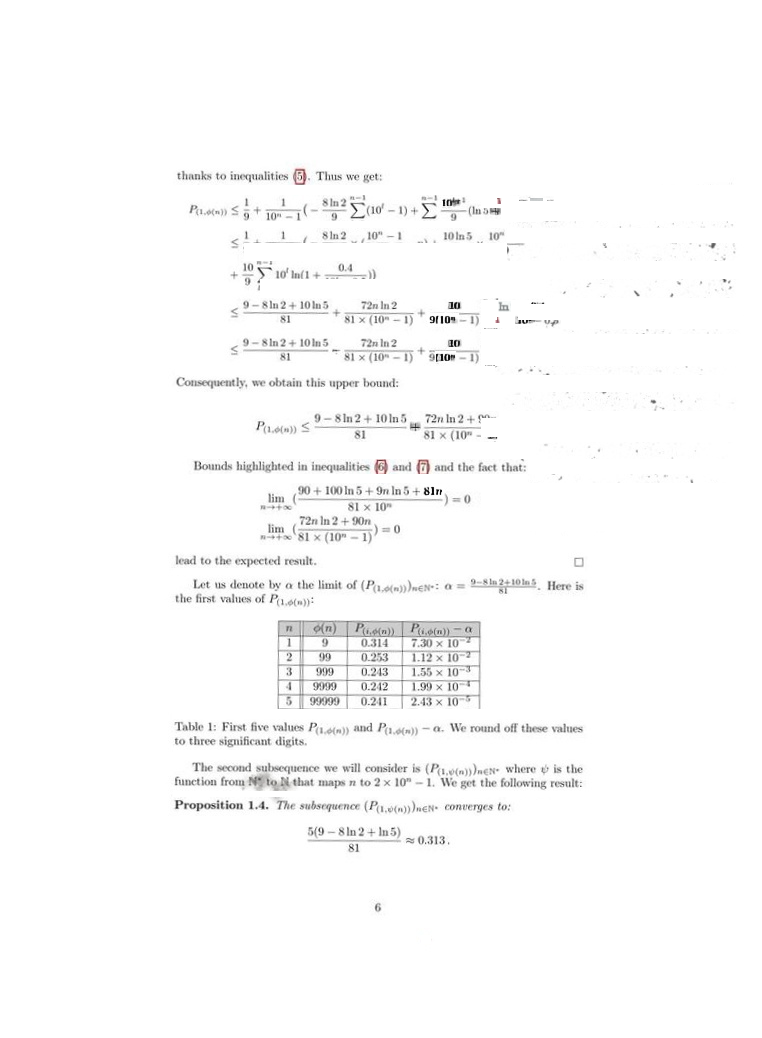} &
        \imgcell{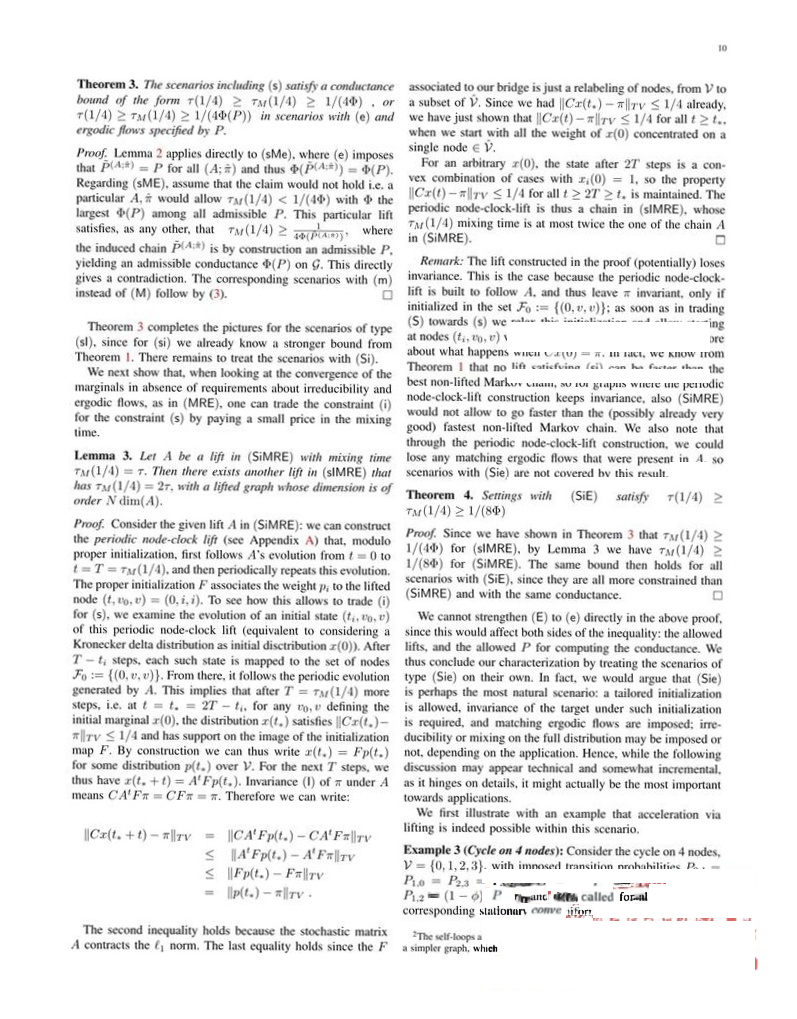} &
        \imgcell{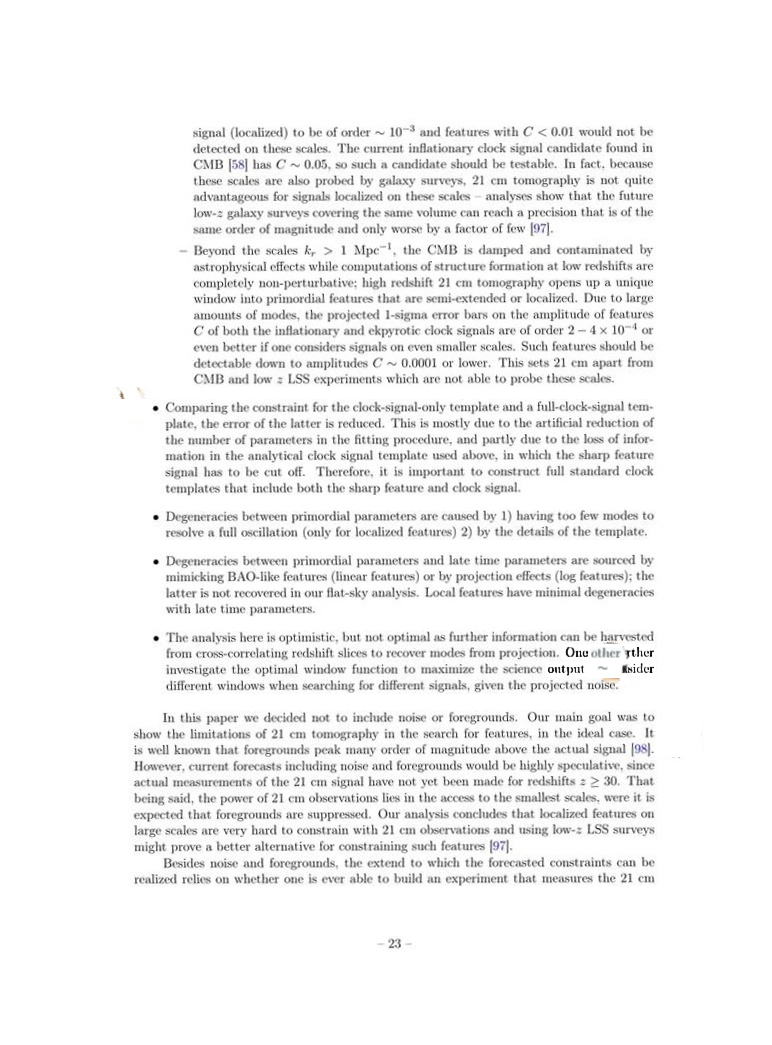} &
        \imgcell{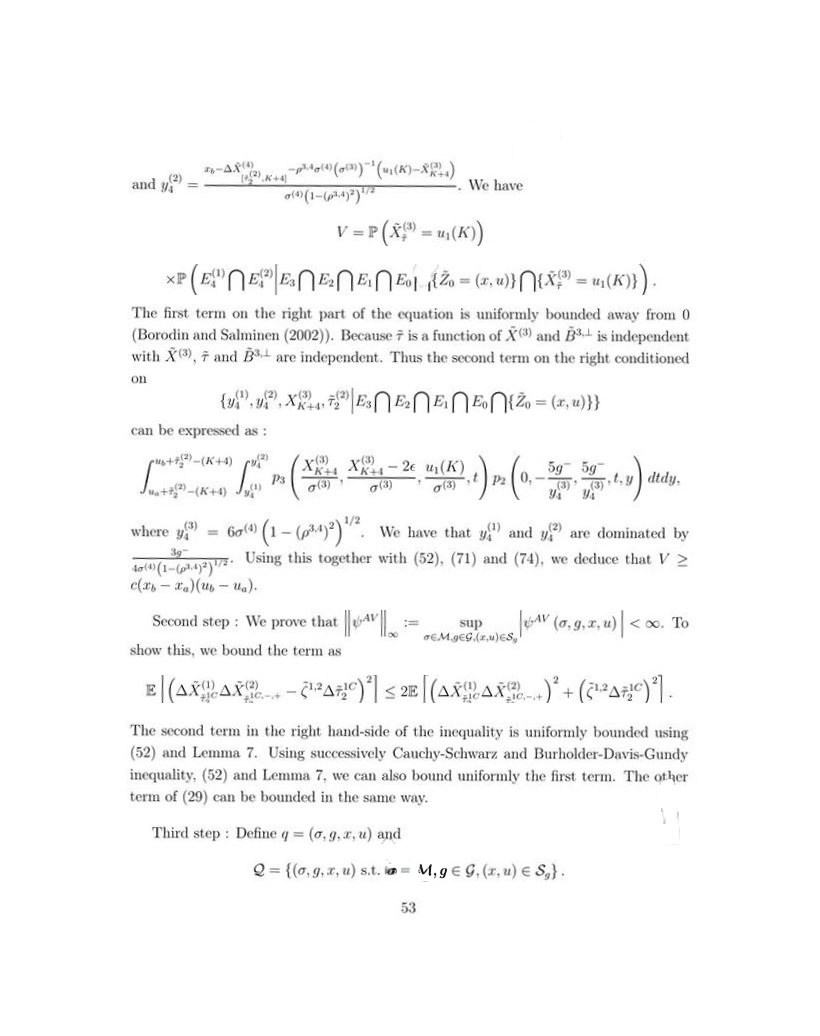} \\

        \rlabel{GT} &
        \imgcell{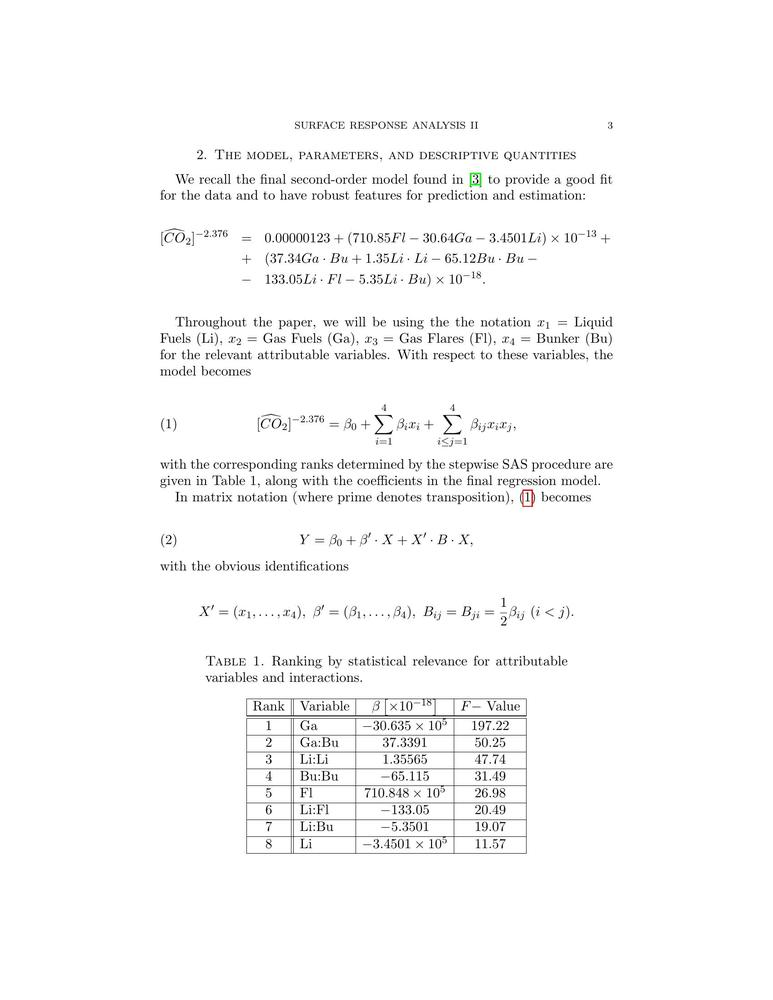} &
        \imgcell{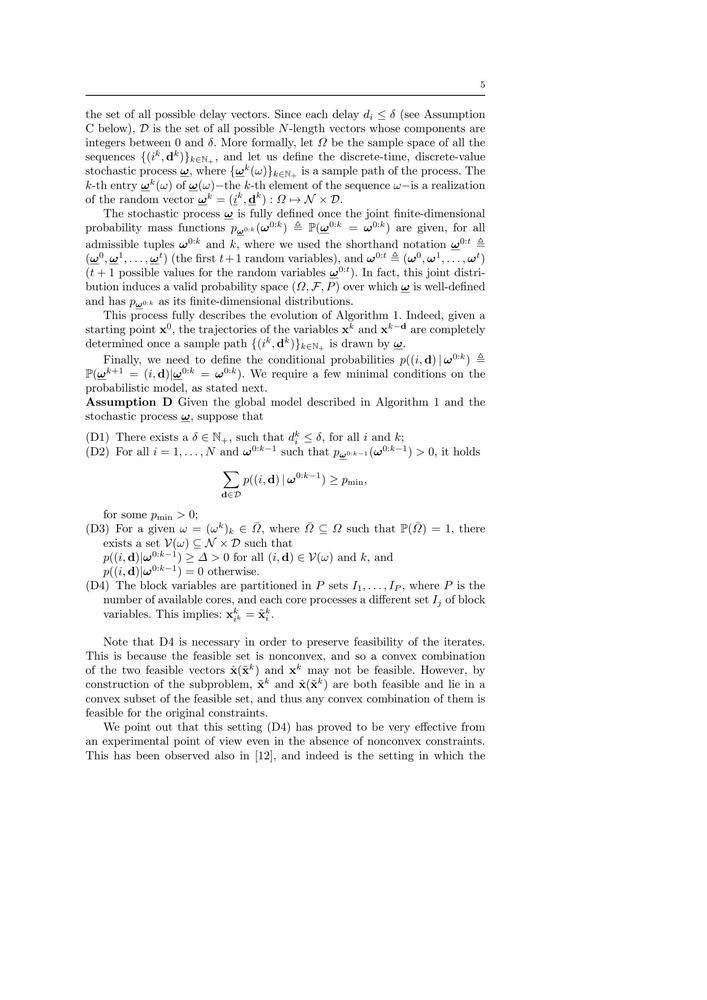} &
        \imgcell{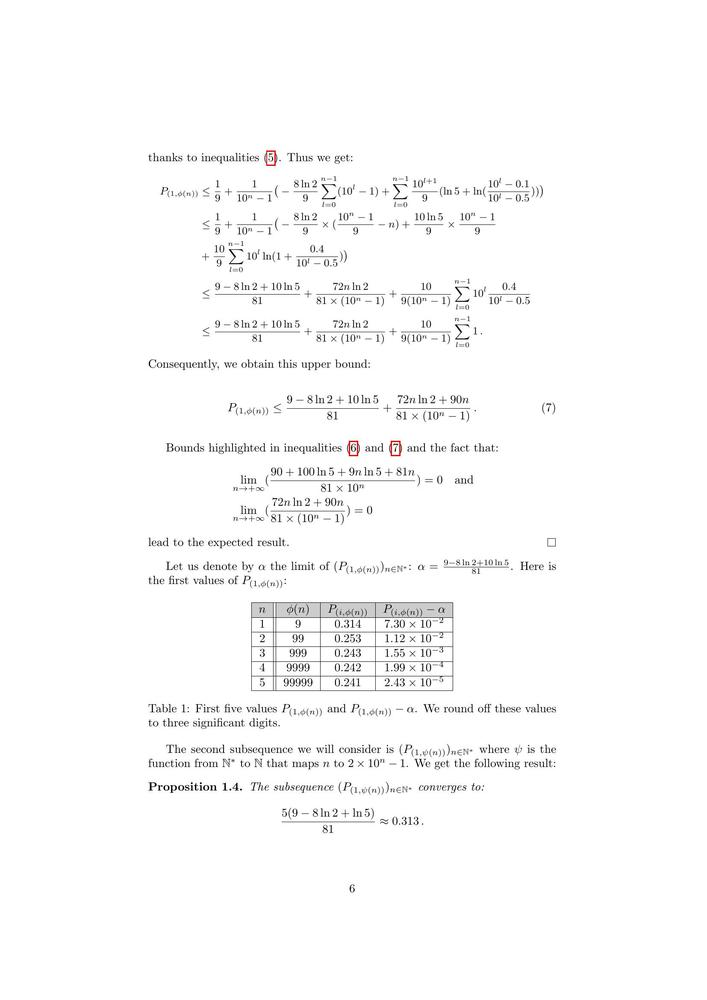} &
        \imgcell{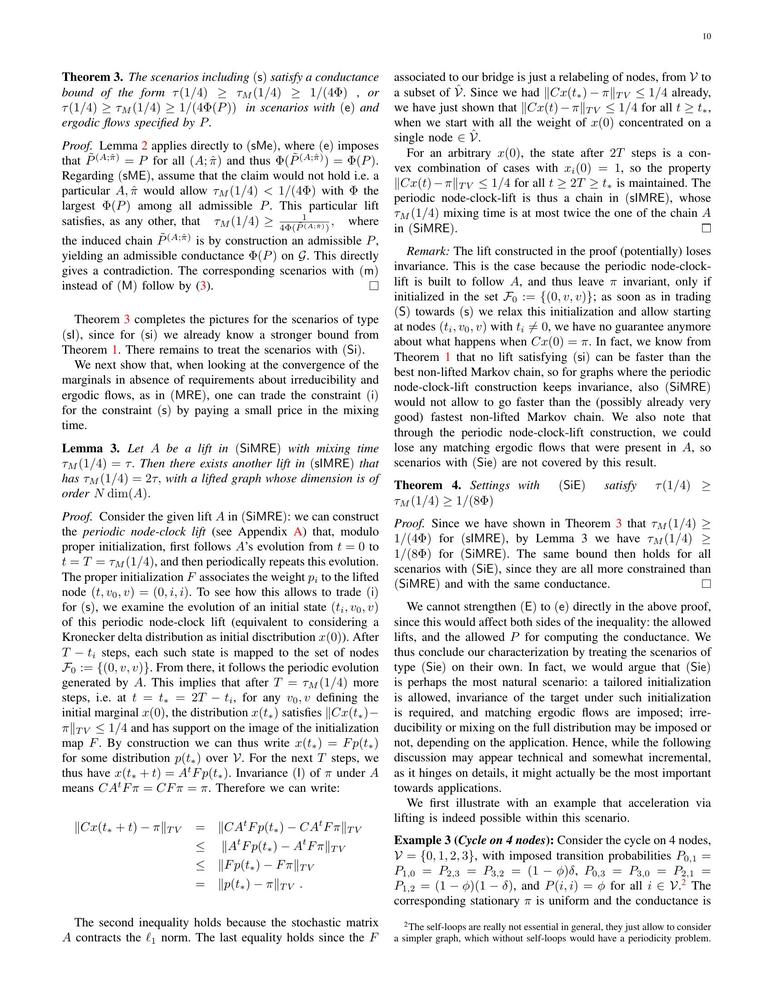} &
        \imgcell{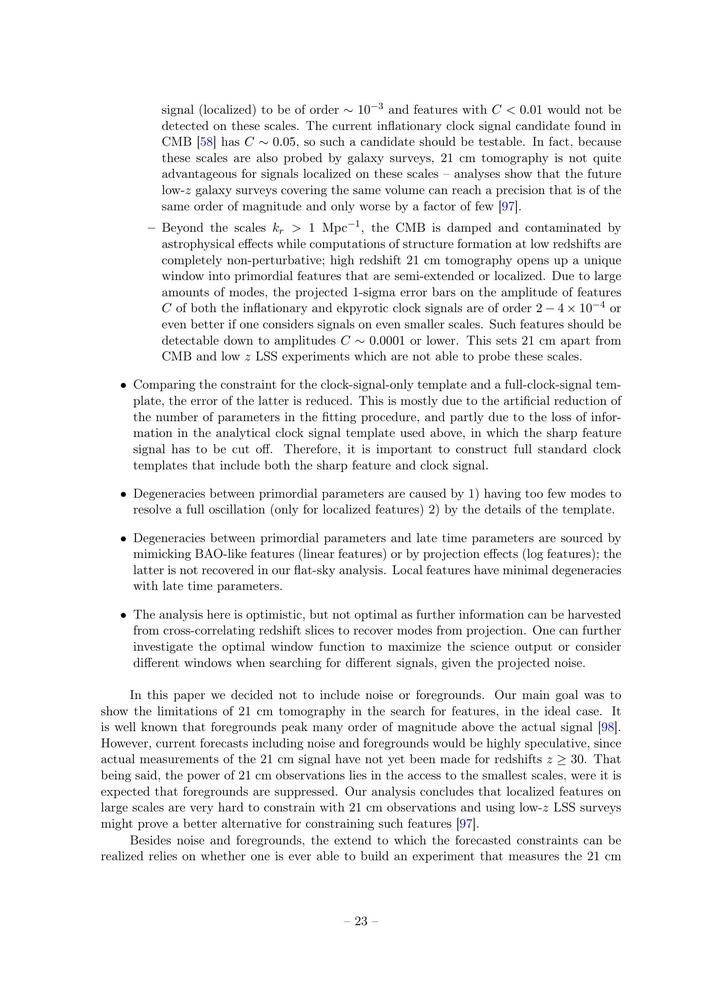} &
        \imgcell{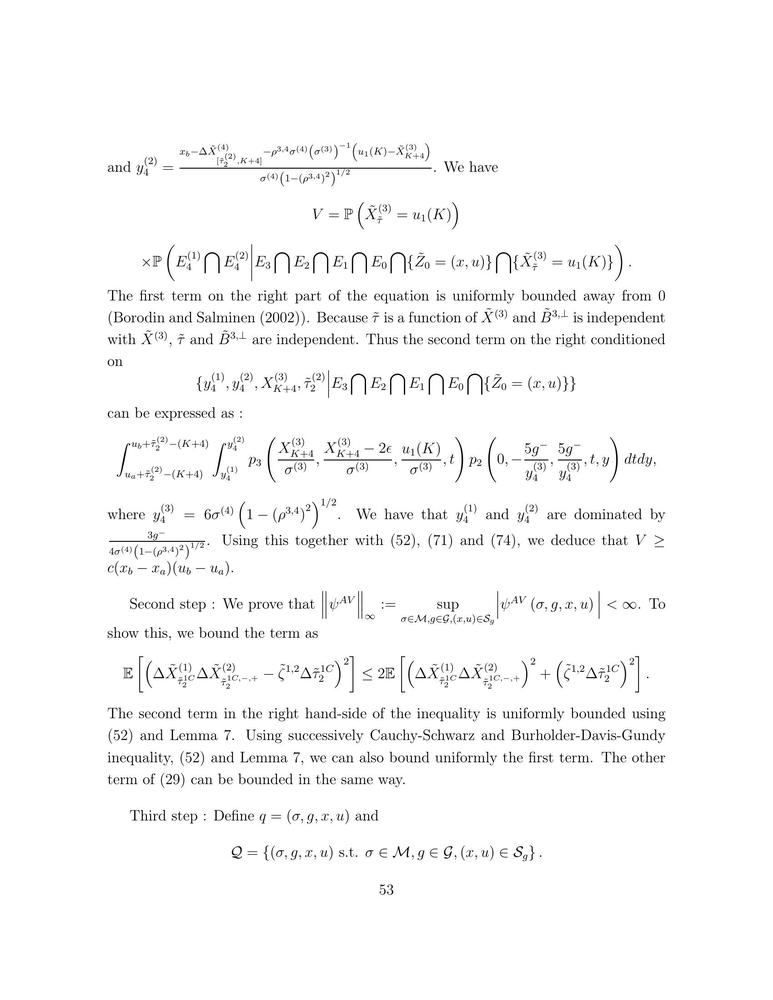} \\

    \end{tabular}

    \caption{Qualitative comparison of document restoration methods across occlusion types.}
    \label{fig:method_comparison}
\end{figure*}

Fig.~\ref{fig:method_comparison} presents a qualitative comparison of DocRevive against NAFNet, DocDiff, and GSDM across six distinct occlusion types. NAFNet, being a general-purpose image restoration network, struggles to handle structured degradations such as ink strokes and stamps, often leaving visible residuals or introducing blurring artifacts in the restored regions. DocDiff, while producing smoother outputs, fails to recover fine typographic details particularly under heavy occlusions like burnt and whitener damage due to its limited conditioning on document structure. GSDM shows improved perceptual quality in certain cases but exhibits hallucinated content in regions with large missing areas, such as scribble and dust degradations. In contrast, DocRevive consistently produces restorations that are both visually coherent and faithful to the ground truth across all occlusion categories, preserving text legibility, line structure, and background consistency. This robustness across diverse degradation types demonstrates the strength of our dataset-driven approach and the model's ability to disentangle occlusion from underlying document content.

\end{document}